\documentclass[]{valhalla}
\usepackage{amsmath,amsfonts,bm}

\def\eqref#1{equation~\ref{#1}}
\def\1{\bm{1}}

\DeclareMathAlphabet{\mathsfit}{\encodingdefault}{\sfdefault}{m}{sl}
\SetMathAlphabet{\mathsfit}{bold}{\encodingdefault}{\sfdefault}{bx}{n}

\usepackage{iftex}
\usepackage{url}
\usepackage{booktabs}
\usepackage{graphicx}
\usepackage{float}
\usepackage{xcolor}
\usepackage{colortbl}
\usepackage{makecell}
\usepackage{pifont}
\usepackage{wrapfig}
\usepackage{hyperref}
\usepackage{array}
\usepackage{multirow}
\usepackage{amsmath}
\usepackage{amssymb}
\usepackage{amsfonts}
\usepackage{nicefrac}
\usepackage{microtype}
\usepackage{longtable}
\usepackage{enumitem}
\usepackage{subcaption}
\usepackage{cleveref}
\usepackage[toc,page,header]{appendix}
\usepackage{minitoc}
\usepackage{tcolorbox}
\tcbuselibrary{most}

\makeatletter
\newcommand{\EpibenchAppendixHeadingSpacing}{%
    \renewcommand{\section}{\@startsection{section}{1}{\z@}{-0.8ex plus -0.2ex minus -.1ex}{0.25ex plus .08ex}{\large\sffamily\bfseries\color{valhallaprimary}\raggedright}}%
    \renewcommand{\subsection}{\@startsection{subsection}{2}{\z@}{-0.55ex plus -0.16ex minus -.08ex}{0.16ex plus .05ex}{\normalsize\sffamily\bfseries\color{valhallaprimary}\raggedright}}%
    \renewcommand{\subsubsection}{\@startsection{subsubsection}{3}{\z@}{-0.18ex plus -0.06ex minus -.04ex}{0.05ex plus .02ex}{\normalsize\sffamily\bfseries\color{valhallaprimary}\raggedright}}%
}
\newcommand{\EpibenchNeedspace}[1]{%
    \par\begingroup
    \dimen@=#1\relax
    \dimen@ii=\pagegoal
    \advance\dimen@ii by -\pagetotal
    \ifdim\dimen@>\dimen@ii
        \pagebreak
    \fi
    \endgroup
}
\makeatother

\newcounter{epialgorithm}
\newcommand{\EpiAlgorithmCaption}[2]{\refstepcounter{epialgorithm}\label{#2}\textbf{Algorithm~\theepialgorithm} #1}
\newcommand{\tick}{\textcolor{valhallaaccent}{\ding{51}}}
\newcommand{\cross}{\textcolor{red!75!black}{\ding{55}}}
\newcommand{\errorhl}[1]{\textcolor{red!75!black}{\textbf{#1}}}
\newcommand{\promptplaceholder}[1]{{\normalfont\$#1\$}}
\definecolor{ReasoningBand}{HTML}{F1EAFE}
\definecolor{NonreasoningBand}{HTML}{F7F3EE}
\newsavebox{\exampleboxcontent}
\newsavebox{\systempromptboxcontent}
\newsavebox{\caseboxcontent}
\newlength{\exampleboxwidth}
\newlength{\systempromptboxwidth}
\newlength{\caseboxwidth}
\newenvironment{examplebox}[1]{%
    \par\vspace{0.75pt}\noindent
    \begingroup
    \def\exampleboxtitle{#1}%
    \setlength{\exampleboxwidth}{0.965\linewidth}%
    \setlength{\fboxsep}{0pt}%
    \begin{lrbox}{\exampleboxcontent}%
    \begin{minipage}{\exampleboxwidth}
    \noindent\colorbox{valhallaprimary}{%
        \makebox[\exampleboxwidth][l]{\hspace{6pt}\color{white}\bfseries\footnotesize\strut \exampleboxtitle}}%
    \par\vspace{5pt}%
    \hspace*{6pt}\begin{minipage}{\dimexpr\exampleboxwidth-12pt\relax}
    \footnotesize\raggedright
}{%
    \end{minipage}\par\vspace{5pt}%
    \end{minipage}%
    \end{lrbox}%
    \setlength{\fboxrule}{0.6pt}%
    \fcolorbox{black!45}{gray!4}{\usebox{\exampleboxcontent}}%
    \endgroup
    \par\vspace{3pt}\needspace{4\baselineskip}
}
\newenvironment{systempromptbox}[1]{%
    \par\vspace{0.75pt}\noindent
    \begingroup
    \def\systempromptboxtitle{#1}%
    \setlength{\systempromptboxwidth}{0.985\linewidth}%
    \setlength{\fboxsep}{0pt}%
    \begin{lrbox}{\systempromptboxcontent}%
    \begin{minipage}{\systempromptboxwidth}
    \noindent\colorbox{valhallaaccent}{%
        \makebox[\systempromptboxwidth][l]{\hspace{8pt}\color{white}\bfseries\normalsize\strut \systempromptboxtitle}}%
    \par\vspace{7pt}%
    \hspace*{8pt}\begin{minipage}{\dimexpr\systempromptboxwidth-16pt\relax}
    \footnotesize\raggedright
}{%
    \end{minipage}\par\vspace{7pt}%
    \end{minipage}%
    \end{lrbox}%
    \setlength{\fboxrule}{0.75pt}%
    \fcolorbox{valhallaaccent}{valhallaaccent!4}{\usebox{\systempromptboxcontent}}%
    \endgroup
    \par\vspace{0.75pt}
}
\newenvironment{casebox}[1]{%
    \par\medskip\noindent
    \begingroup
    \def\caseboxtitle{#1}%
    \setlength{\caseboxwidth}{0.985\linewidth}%
    \setlength{\fboxsep}{0pt}%
    \begin{lrbox}{\caseboxcontent}%
    \begin{minipage}{\caseboxwidth}
    \noindent\colorbox{valhallasecondary}{%
        \makebox[\caseboxwidth][l]{\hspace{8pt}\color{white}\bfseries\normalsize\strut \caseboxtitle}}%
    \par\vspace{7pt}%
    \hspace*{8pt}\begin{minipage}{\dimexpr\caseboxwidth-16pt\relax}
    \footnotesize\raggedright
}{%
    \end{minipage}\par\vspace{7pt}%
    \end{minipage}%
    \end{lrbox}%
    \setlength{\fboxrule}{0.75pt}%
    \fcolorbox{valhallasecondary}{valhallasecondary!4}{\usebox{\caseboxcontent}}%
    \endgroup
    \par\medskip
}

\paperlogo{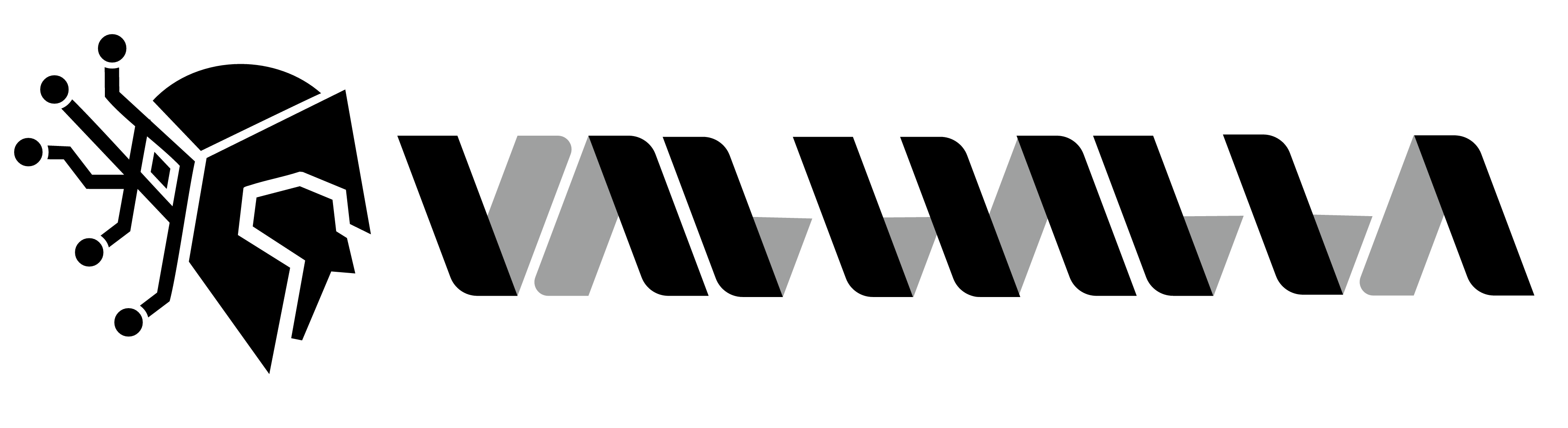}
\papertagline{}

\title{EpiBench: Can LLMs Understand Epitopes for Antibody Drug Discovery?}

\author[1, *]{Zirui Wang}
\author[1, *]{Jiaqi Wang}
\author[1]{Qinghan Wang}
\author[1]{Yuzhi Xu}
\author[1]{Gang Du}
\author[2]{Tingjun Hou}
\author[1,\dagger]{Odin Zhang}

\affiliation[1]{Valhalla Technology}
\affiliation[2]{Zhejiang University}

\contribution[*]{Core Contributors}
\contribution[\dagger]{Corresponding Author}

\abstract{Epitopes determine where antibodies bind antigens and shape downstream therapeutic properties such as functional blockade and escape resistance, making epitope understanding central to antibody drug discovery. Although large language models (LLMs) have shown strong biomedical reasoning ability, it remains unclear whether they can infer epitope information directly from antigen and antibody sequences. Existing epitope resources typically focus on isolated prediction tasks or rely on specialized structural settings, while general protein benchmarks do not evaluate epitope-centered decisions across the antibody development workflow. To address this gap, we introduce EpiBench, a closed-book, sequence-based, and automatically scorable benchmark for evaluating epitope reasoning in LLMs. EpiBench contains 1,609 curated samples grounded in structural antibody--antigen contacts, curated functional B-cell assays, and deep mutational scanning escape measurements. It covers five connected tasks: targetable region discovery, antibody-conditioned epitope identification, epitope binning, functional epitope assessment, and antibody escape assessment, with controlled sampling to reduce shortcut-based evaluation artifacts. We evaluate nine general-purpose LLMs and analyze their behavior through task-specific baselines, antigen length stratification, explicit-reasoning comparison, and failure-mode inspection. The results show that current LLMs capture partial epitope-related signals but remain limited in antibody-specific sequence grounding, long-context residue localization, and biologically grounded reasoning. Therefore, EpiBench provides a diagnostic testbed for measuring and improving sequence-aware biomedical LLMs toward reliable LLM-assisted antibody discovery.}

\correspondence{odin@valtech.ai}
\checkdata[Data]{\url{https://huggingface.co/datasets/oteam/EpiBench}}
\begin{document}
\maketitle

\begin{figure}[t]
    \centering
    \includegraphics[width=\linewidth]{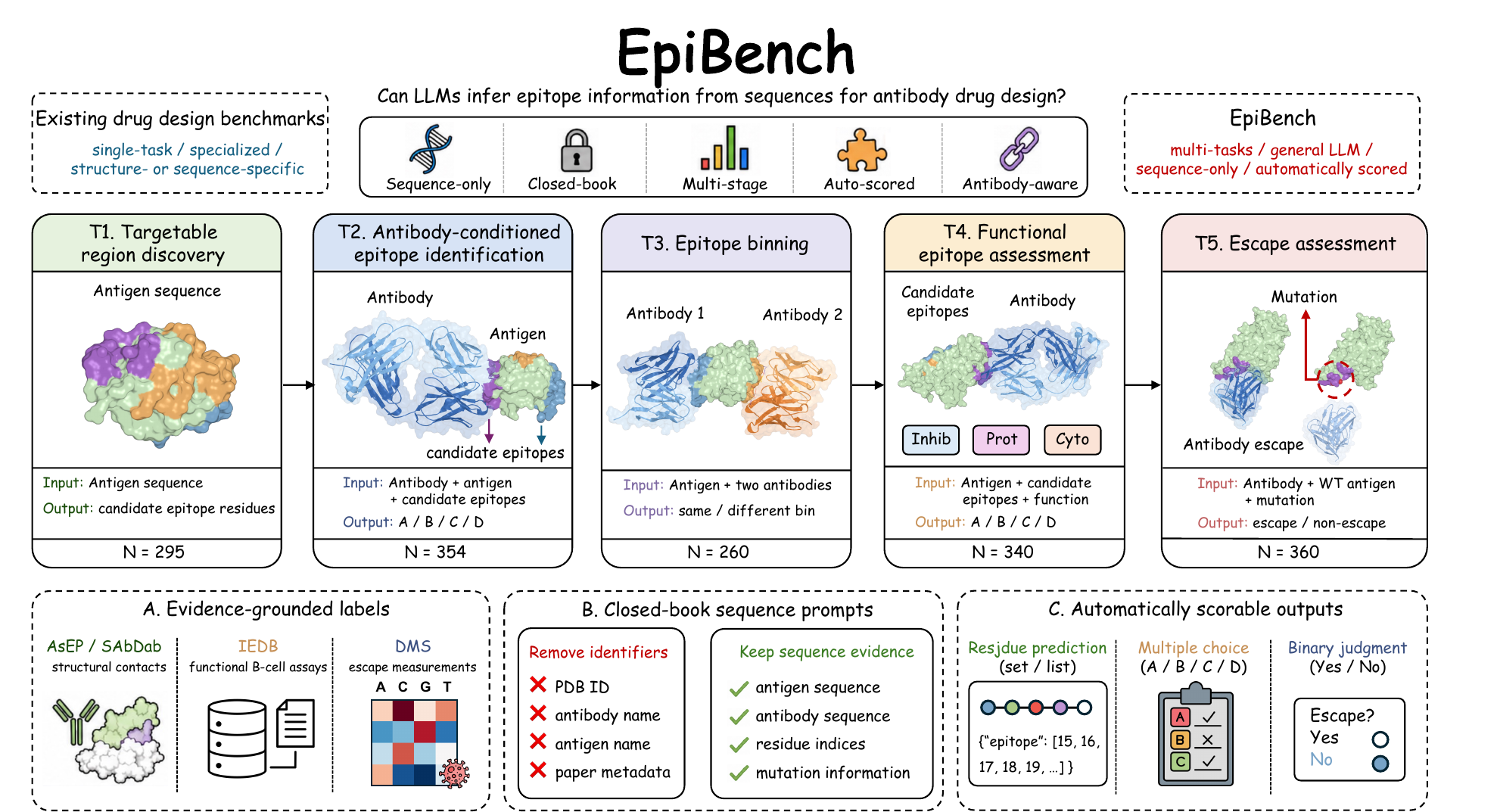}
    \caption{Overview of EpiBench. The benchmark evaluates whether LLMs can infer epitope information from antigen and antibody sequences through a closed-book, sequence-based pipeline covering targetable region discovery, antibody-conditioned epitope identification, epitope binning, functional epitope assessment, and escape assessment. Labels are grounded in structural contacts, curated functional assays, and deep mutational scanning evidence, while prompts remove identifiers and retain sequence-level evidence for automatically scorable evaluation.}
    \label{fig:epibench-overview}
\end{figure}

\section{Introduction}

The discovery of therapeutic antibodies is central to modern biomedicine and increasingly important for precision treatment across diverse diseases. The therapeutic effect of an antibody depends not only on whether it recognizes an antigen but also on which antigen region it engages. This binding region is known as the epitope, and it shapes target engagement, functional blockade, immune escape, and therapeutic specificity. Understanding antibody antigen binding through epitopes is therefore a fundamental problem in antibody drug design and optimization. In recent years, task specific deep learning methods have been developed to predict epitopes from antigen sequences or protein structures \citep{clifford2022bepipred,ponomarenko2008ellipro,liu2024asep}. These specialized models provide evidence that epitope related signals are computationally predictable, yet they are usually optimized for a narrow prediction target and do not directly address broader epitope centered decisions in antibody drug development. In parallel, protein language models such as ESM have expanded protein sequence modeling toward general representation learning, showing that large scale self supervised training can capture structural and functional regularities from biological sequences \citep{rives2021biological,lin2023evolutionary,wang2023unirna}. Together, these advances suggest that antigen and antibody sequences contain learnable signals for reasoning about antibody antigen interfaces.

Meanwhile, LLMs have demonstrated strong understanding and reasoning abilities across biomedical tasks \citep{singhal2022large}, but their ability to infer epitope information from antigen and antibody sequences remains unclear. This motivates a central question: \textbf{Can LLMs infer epitope information from antigen and antibody sequences and use it to support antibody drug design?} Addressing this question requires moving beyond isolated epitope prediction and evaluating whether LLMs can support a sequence driven decision chain that includes binding site localization, antibody grouping, functional interpretation, and escape assessment.

To answer this question, we introduce EpiBench, a sequence based benchmark for evaluating epitope reasoning in antibody drug discovery. EpiBench is designed to simulate the workflow of antibody drug development, where epitope information supports multiple decisions across discovery, characterization, and optimization. It covers five connected decision tasks, including antibody free epitope prediction, antibody conditioned epitope identification, antibody epitope binning, functional epitope assessment, and antibody escape assessment. The benchmark contains 1,609 curated samples constructed with controlled sampling strategies to improve diversity, balance answer distributions, and reduce shortcut based evaluation artifacts. According to the biological objective of each task, we formulate the samples as question answering, multiple choice, and binary judgment problems, and design task adapted evaluation metrics for systematic assessment. We summarize our main contributions below.

\begin{itemize}
    \item We introduce EpiBench, a standardized benchmark for evaluating LLMs on epitope reasoning in antibody drug discovery. EpiBench is organized around the practical workflow of antibody development and covers five key stages spanning epitope localization, antibody conditioned recognition, epitope binning, functional assessment, and escape analysis.
    \item We conduct a comprehensive evaluation of nine LLMs and compare them with task-specific epitope models where directly applicable. The results show that current LLMs capture partial epitope-related signals, but remain limited in long-context residue localization, antibody-specific sequence grounding, and tasks requiring indirect functional or escape reasoning.
    \item We provide a fine grained analysis of model behavior across tasks, revealing systematic failure modes in antibody antigen interface reasoning. These findings clarify current limitations of LLMs for epitope centered drug discovery and suggest directions for future biological sequence reasoning models.
\end{itemize}

\section{Related Work}

\subsection{Task Specific Epitope Prediction Models}

Computational prediction of antibody antigen interfaces has traditionally been studied through task specific epitope prediction models. Early sequence based methods predicted linear B cell epitopes from antigen primary sequences using physicochemical propensity scores, local sequence windows, recurrent neural networks, and kernel based classifiers \citep{emini1985induction,kolaskar1990semi,larsen2006improved,saha2006prediction,el2008predicting}. More recent methods further incorporate deep learning and protein language representations to improve epitope prediction under antigen only or antibody conditioned settings \citep{clifford2022bepipred,liu2024asep}. These studies provide direct evidence that epitope related signals can be learned from biological sequences, but their learning objectives are usually tied to a single prediction target and a fixed input format.

Complementary studies predict conformational epitopes by using three dimensional antigen structures, solvent exposure, surface geometry, residue neighborhoods, or sequence derived approximations of spatial binding sites \citep{haste2006prediction,ponomarenko2008ellipro,sun2009seppa,ansari2010identification,wu2024fafe,xu2026sakepp}. However, most existing methods are designed around a single prediction endpoint, which leaves multi task epitope reasoning in antibody drug design insufficiently examined. EpiBench addresses this evaluation gap by organizing the assessment around five key stages, including epitope localization, antibody conditioned recognition, epitope binning, functional assessment, and escape analysis.

\subsection{Protein Language Models and Sequence Understanding Benchmarks}

Protein language models have reframed protein sequence analysis as self supervised representation learning over amino acid sequences. Models such as ProtTrans and ESM train Transformer architectures on large protein corpora and show that learned representations encode structural, functional, and evolutionary information \citep{elnaggar2021others,rives2021biological}. ESMFold further demonstrates sequence to structure prediction from language model representations, while autoregressive models such as ProGen and ProGen2 extend language modeling to controllable protein sequence generation \citep{lin2023evolutionary,madani2023large,nijkamp2023progen2,li2025pephar}. Recent biomolecular LLM studies further connect biological sequences with textual instructions, as shown by Mol-Instructions and ProtLLM \citep{fang2024mol,zhuo2024protllm}. Together, these works support the view that protein sequences contain rich learnable signals that can be exploited by language modeling.

Protein sequence benchmarks have provided standardized settings for measuring these modeling capabilities. TAPE evaluates transfer learning across structural and functional protein tasks, PEER expands protein sequence understanding to function, localization, structure, protein protein interaction, and protein ligand interaction prediction, and fitness oriented benchmarks such as FLIP and ProteinGym evaluate few shot fitness prediction and variant effect modeling \citep{rao2019evaluating,xu2022peer,mollon2025exploring,notin2022tranception,guo2024retrieval}. These benchmarks are essential for assessing general protein sequence representations, but they primarily focus on broad biological properties, fitness landscapes, or variant effects. They do not directly examine epitope centered reasoning for antibody antigen recognition, where models must relate binding site localization, antibody grouping, functional interpretation, and escape assessment within a drug development context. EpiBench focuses on this missing setting by evaluating epitope reasoning across tasks aligned with antibody drug discovery.

\begin{table}[t]
    \vspace{-2mm}
    \caption{Comparison with representative epitope, antibody, and protein benchmarks shows that EpiBench uniquely combines sequence-accessible epitope-centered antibody--antigen reasoning, multi-source evidence, drug-discovery objectives, LLM-ready formulation, and shortcut-controlled automatic evaluation.}
    \label{tab:benchmark-comparison}
    \centering
    \scriptsize
    \setlength{\tabcolsep}{3.2pt}
    \renewcommand{\arraystretch}{1.08}
    \resizebox{\textwidth}{!}{%
    \begin{tabular}{l|cccc|cc|cccc}
        \toprule
        &
        \multicolumn{4}{c|}{\textbf{Dataset}} &
        \multicolumn{2}{c|}{\textbf{Task}} &
        \multicolumn{4}{c}{\textbf{Evaluation}} \\
        \cline{2-11}
        &
        \textbf{\makecell{Epitope\\centered}} &
        \textbf{\makecell{Ab--Ag\\context}} &
        \textbf{\makecell{Sequence\\input}} &
        \textbf{\makecell{Multi-source\\evidence}} &
        \textbf{\makecell{Multi-\\objective}} &
        \textbf{\makecell{Drug\\discovery}} &
        \textbf{\makecell{LLM-\\ready}} &
        \textbf{\makecell{Standard\\protocol}} &
        \textbf{\makecell{Auto\\scoring}} &
        \textbf{\makecell{Bias\\control}} \\
        \midrule
        ABCPred~\citep{saha2006prediction} & \tick & \cross & \tick & \cross & \cross & \cross & \cross & \tick & \tick & \cross \\
        BCPred~\citep{el2008predicting} & \tick & \cross & \tick & \cross & \cross & \cross & \cross & \tick & \tick & \cross \\
        BepiPred-3.0 eval set~\citep{clifford2022bepipred} & \tick & \cross & \tick & \tick & \cross & \cross & \cross & \tick & \tick & \cross \\
        AsEP~\citep{liu2024asep} & \tick & \tick & \cross & \cross & \cross & \tick & \cross & \tick & \tick & \tick \\
        CHIMERA-Bench~\citep{ahmed2026chimera} & \tick & \tick & \cross & \cross & \tick & \tick & \cross & \tick & \tick & \tick \\
        TDC immune tasks~\citep{huang2022artificial} & \tick & \cross & \tick & \tick & \tick & \tick & \cross & \tick & \tick & \cross \\
        TDC developability~\citep{huang2022artificial} & \cross & \cross & \tick & \cross & \cross & \tick & \cross & \tick & \tick & \cross \\
        Ginkgo AbDev~\citep{van20262025} & \cross & \cross & \tick & \cross & \tick & \tick & \cross & \tick & \tick & \tick \\
        FLAb~\citep{chungyoun2024flab} & \cross & \cross & \tick & \tick & \tick & \tick & \cross & \tick & \tick & \cross \\
        AbBiBench~\citep{zhao2025benchmark} & \cross & \tick & \tick & \tick & \tick & \tick & \cross & \tick & \tick & \tick \\
        TAPE~\citep{rao2019evaluating} & \cross & \cross & \tick & \tick & \tick & \cross & \cross & \tick & \tick & \tick \\
        PEER~\citep{xu2022peer} & \cross & \cross & \tick & \tick & \tick & \cross & \cross & \tick & \tick & \tick \\
        ProteinGLUE~\citep{capel2022proteinglue} & \cross & \cross & \tick & \tick & \tick & \cross & \cross & \tick & \tick & \cross \\
        PETA~\citep{tan2024peta} & \cross & \cross & \tick & \tick & \tick & \cross & \cross & \tick & \tick & \tick \\
        ProteinGym~\citep{notin2023proteingym} & \cross & \cross & \tick & \tick & \tick & \tick & \cross & \tick & \tick & \tick \\
        FLIP/FLIP2~\citep{dallago2021flip,didi2026flip2} & \cross & \cross & \tick & \tick & \tick & \tick & \cross & \tick & \tick & \tick \\
        ATOM3D~\citep{townshend2021atom3d} & \cross & \cross & \cross & \tick & \tick & \tick & \cross & \tick & \tick & \tick \\
        \rowcolor{blue!10}
        EpiBench (Ours) & \tick & \tick & \tick & \tick & \tick & \tick & \tick & \tick & \tick & \tick \\
        \bottomrule
    \end{tabular}%
    }
    \vspace{-2mm}
\end{table}

\section{EpiBench}

\subsection{Design Principles}

Biomedical LLM evaluation has largely centered on textual scientific knowledge, including biomedical question answering, literature search, evidence retrieval, and experimental report interpretation. These settings are important for measuring how models access and organize biomedical information, but they do not directly assess whether language models can reason from biological sequences. EpiBench complements this line of evaluation by constructing a sequence centered benchmark in which models must infer epitope related information from antigen and antibody sequences and use it in molecular decision tasks. This setting directly links scientific reasoning with biological sequence understanding and provides a controlled testbed for examining the potential of general purpose LLMs in complex sequence driven scientific problems.

The construction of EpiBench follows five principles: (1) aligning tasks with key stages of antibody development, from candidate epitope discovery to escape assessment; (2) deriving labels from experimentally grounded sources, including structural antibody antigen complexes, curated functional B cell assays, and deep mutational scanning measurements \citep{liu2024asep,dunbar2014sabdab,vita2025immune,cao2022ba}; (3) using a closed book sequence formulation that removes identifiers such as PDB IDs, antigen names, antibody names, accession numbers, native numbering schemes, and publication metadata; (4) adopting task adapted answer formats, including residue prediction, multiple choice, and binary judgment; and (5) controlling answer distributions, candidate orders, distractors, and matched pairs to reduce shortcut based evaluation artifacts.

\subsection{Benchmark Task Design}

EpiBench converts epitope centered decisions in antibody drug discovery into task specific prompts with automatically scorable outputs. This section describes the motivation and high level construction strategy for each task, while detailed data filtering, label construction, and sampling procedures are provided in Appendix B.1.

\textbf{Task1: Antibody Free Epitope Prediction.} This task evaluates whether a model can identify antigen regions that are likely to be recognized before a specific antibody candidate is available. This setting reflects early target and epitope assessment, where researchers need to prioritize potentially targetable regions on an antigen sequence. We operationalize the task by asking the model to predict a limited set of candidate epitope residues from a position numbered antigen sequence. The reference label is constructed from the union of structurally observed antibody contact residues on the same antigen, which supports recall oriented evaluation while acknowledging that known epitopes may be incomplete \citep{liu2024asep,dunbar2014sabdab}.

\textbf{Task2: Antibody Conditioned Epitope Identification.} This task evaluates whether a model can associate a specific antibody sequence with its corresponding binding region on an antigen. This task reflects mechanism characterization for candidate antibodies, where the key question is not whether a region is generally epitope like, but which region is recognized by the given antibody. We formulate the task as a multiple choice problem over candidate epitopes, using the true structural epitope as the answer and constructing distractors from other real epitopes on the same antigen or matched synthetic regions. This design keeps the task automatically scorable while preserving antibody specific reasoning.

\textbf{Task3: Antibody Epitope Binning.} This task evaluates whether a model can determine whether two antibodies bind overlapping antigen regions. This task corresponds to antibody panel organization, where epitope redundancy and diversity affect lead selection, combination design, and downstream characterization. We formulate binning as a binary judgment over two antibodies against the same antigen, with labels derived from structural epitope overlap. To reduce shortcuts based only on antibody sequence similarity, positive and negative pairs are sampled with matched antibody similarity profiles when possible.

\textbf{Task4: Functional Epitope Assessment.} This task evaluates whether a model can distinguish epitopes associated with functional antibody outcomes from epitopes that only show binding evidence or lack functional support. This task reflects the distinction between antigen recognition and therapeutic relevance, since not every binding site leads to neutralization, inhibition, protection, or effector mediated activity. We formulate the task as a multiple choice problem over same antigen candidate epitopes, where the answer is supported by curated functional B cell assay evidence and distractors are selected from binding only or property matched candidate regions \citep{vita2025immune}. This design encourages models to reason about functional consequence rather than epitope likelihood alone.

\textbf{Task5: Antibody Escape Assessment.} This task evaluates whether an antigen point mutation disrupts recognition by a particular antibody. This task reflects lead optimization and resistance risk assessment, where the same mutation may have different effects across antibodies depending on their binding interfaces. We formulate escape assessment as a binary judgment from antibody variable domain sequences, a wild type antigen sequence, and a single antigen mutation. Labels are derived from deep mutational scanning measurements, and mutation level matching is used to reduce shortcuts based only on mutation identity \citep{cao2022ba}. The prompt removes antigen and antibody identifiers so that the task emphasizes sequence conditioned escape reasoning.

\subsection{Dataset Statistics}

\begin{figure}[t]
    \centering
    \includegraphics[width=0.86\linewidth]{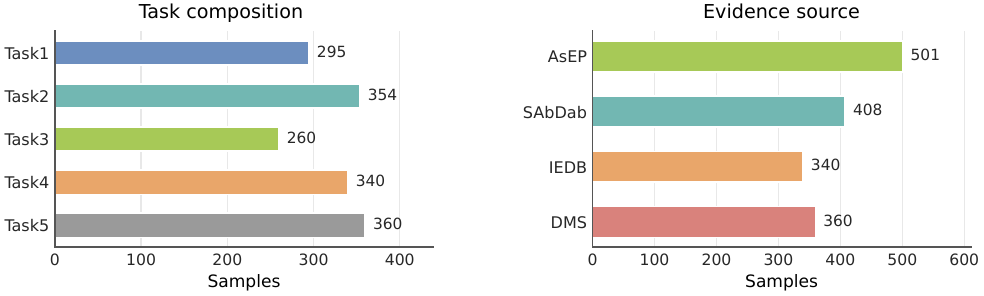}
    \caption{Dataset composition of EpiBench. The two panels report sample counts across the five benchmark tasks and four evidence sources. T1--T5 denote targetable region discovery, antibody-conditioned epitope identification, epitope binning, functional epitope assessment, and antibody escape assessment.}
    \label{fig:dataset-composition}
\end{figure}

\begin{figure}[t]
    \centering
    \includegraphics[width=0.92\linewidth]{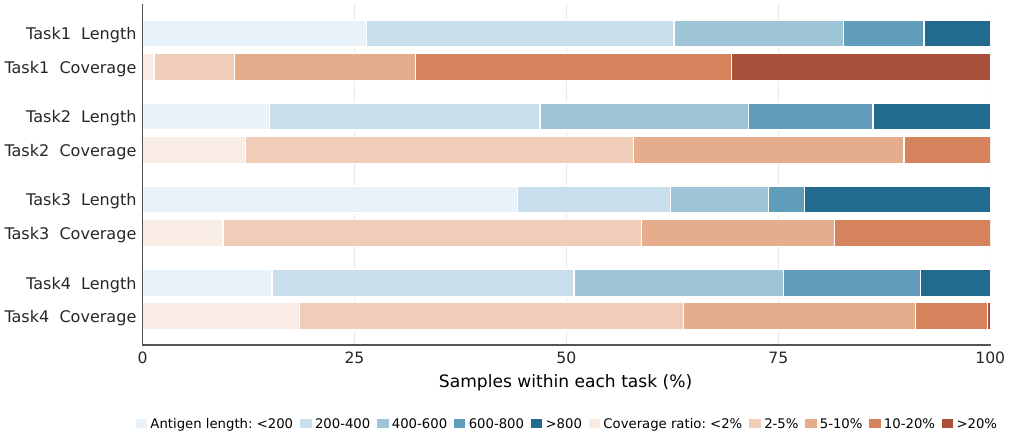}
    \caption{Antigen length and epitope coverage distributions in EpiBench. For each of Tasks~1--4, the upper stacked bar shows the antigen length distribution and the lower stacked bar shows the epitope coverage ratio distribution. Task~5 is excluded because it evaluates mutation-specific escape on a fixed antigen window and does not define an explicit epitope residue set for coverage analysis.}
    \label{fig:dataset-sequence-distribution}
\end{figure}

Figures~\ref{fig:dataset-composition} and~\ref{fig:dataset-sequence-distribution} summarize the composition and sequence scale of EpiBench. The benchmark contains 1,609 samples across five tasks, with each task contributing between 15\% and 25\% of the full dataset. The samples are derived from four source datasets, including AsEP \citep{liu2024asep}, SAbDab \citep{dunbar2014sabdab}, DMS escape measurements \citep{cao2022ba}, and IEDB \citep{vita2025immune}. We report sequence length and epitope coverage statistics only for Tasks 1 to 4 because Task5 is designed for mutation specific escape assessment, does not require full length antigen input, and does not define an explicit epitope residue set for coverage analysis. Overall, EpiBench includes a representative collection of antigen sequences and task relevant evidence, enabling systematic evaluation of LLMs on epitope reasoning.

\section{Experiments}

\subsection{Overall Model Performance}

Table~\ref{tab:main-results} reports the performance of nine general-purpose LLMs across the five EpiBench tasks. We select models according to two dimensions, weight accessibility and reasoning capability. The resulting set covers closed-source reasoning models (Gemini-3-Flash, GPT-5.5, and Qwen3.7-Plus), closed-source nonreasoning models (GPT-4o and Grok 4.20), open-weight reasoning models (GLM-5.2, DeepSeek-V4-Pro, and Kimi-K2.6), and an open-weight nonreasoning model (Qwen3-235B). All models are evaluated under a unified zero-shot protocol that permits explicit intermediate reasoning. Detailed model and prompt configurations are provided in Appendix~C. To provide additional context, the bottom rows of Table~\ref{tab:main-results} include random baselines and specialized epitope models where directly applicable. The specialized comparison uses BepiPred-3.0 for Task~1 and EpiPred-derived baselines for Tasks~2--3, while Tasks~4--5 are marked as N/A because we did not identify established sequence-based models that directly match EpiBench's functional epitope assessment and antibody-specific escape settings. Based on these evaluations, we make three observations.

\begin{table}[t]
    \centering
    \caption{Overall zero-shot performance of nine general-purpose LLMs on EpiBench (0--100 scale), with random and specialized-model references shown at the bottom. Task~1 reports RegionRecall@50 (RegR@50) and AUROC, while Tasks~2--5 report accuracy. Within LLM rows, \textbf{bold} and \underline{underlined} values denote the best and second-best results, respectively. N/A indicates that the corresponding reference score is not applicable or no directly matched specialized model is available.}
    \label{tab:main-results}
    \resizebox{\textwidth}{!}{%
    \begin{tabular}{lcccccc}
        \toprule
        &
        \multicolumn{2}{c}{\textbf{\makecell{T1: Region Discovery}}} &
        \textbf{\makecell{T2: Epitope\\Identification}} &
        \textbf{\makecell{T3: Epitope\\Binning}} &
        \textbf{\makecell{T4: Functional\\Assessment}} &
        \textbf{\makecell{T5: Escape\\Risk}} \\
        \cmidrule(lr){2-3}\cmidrule(lr){4-4}\cmidrule(lr){5-5}\cmidrule(lr){6-6}\cmidrule(lr){7-7}
        & \textbf{RegR@50 $\uparrow$} & \textbf{AUROC $\uparrow$}
        & \textbf{Acc. $\uparrow$} & \textbf{Acc. $\uparrow$}
        & \textbf{Acc. $\uparrow$} & \textbf{Acc. $\uparrow$} \\
        \rowcolor{ReasoningBand}
        \multicolumn{7}{c}{\textit{Reasoning Models}} \\
        Gemini-3-Flash & \textbf{56.9} & \textbf{54.6} & 28.2 & 53.5 & \textbf{35.6} & 46.2 \\
        GPT-5.5 & \underline{49.7} & \underline{52.3} & \underline{31.9} & \textbf{67.3} & \underline{33.2} & \underline{52.2} \\
        Qwen3.7-Plus & 47.8 & 52.0 & 24.0 & 58.9 & 25.0 & 48.6 \\
        GLM-5.2 & 45.0 & 51.4 & 24.3 & 56.9 & 24.1 & 50.0 \\
        DeepSeek-V4-Pro & 31.8 & 51.3 & 30.8 & 55.0 & 30.3 & 48.3 \\
        Kimi-K2.6 & 36.7 & 51.1 & 25.4 & 51.9 & 27.1 & 41.4 \\
        \rowcolor{NonreasoningBand}
        \multicolumn{7}{c}{\textit{Nonreasoning Models}} \\
        GPT-4o & 38.5 & 50.5 & 25.4 & 51.2 & 23.8 & \underline{52.2} \\
        Grok 4.20 & 40.8 & 50.9 & \textbf{33.6} & \underline{65.8} & 30.0 & \textbf{54.7} \\
        Qwen3-235B & 40.5 & 50.3 & 29.4 & 49.2 & 24.4 & 46.7 \\
        \midrule
        Random baseline & \textit{N/A} & \textit{50.0} & \textit{25.0} & \textit{50.0} & \textit{25.0} & \textit{50.0} \\
        Specialized model & 60.4 & 58.0 & 40.2 & 70.8 & \textit{N/A} & \textit{N/A} \\
        \bottomrule
    \end{tabular}%
    }
\end{table}

\noindent\textbf{(1) Current LLMs capture coarse epitope signals but remain weak at precise interface discrimination.} On Task~1, the clear improvement in RegionRecall@50 shows that some models can prioritize broadly relevant antigen regions, whereas AUROC remains close to random, indicating that they do not reliably distinguish epitope from non-epitope residues across the sequence. Performance is similarly limited on Tasks~2 and 5, where success requires conditioning the prediction on a particular antibody and resolving antibody-specific binding or escape effects. These results suggest that recognizing generic antigen properties is more accessible than grounding a specific antibody--antigen interaction from sequence alone.

\noindent\textbf{(2) Task formulation and biological dependency strongly shape the apparent difficulty.} Task~3 shows the largest gain over its random baseline, but it asks for a binary comparison between two antibodies and may admit pairwise sequence-similarity cues without requiring explicit localization of the binding interface. By contrast, Task~4 requires connecting an epitope to downstream functional evidence, and Task~5 further requires predicting the effect of a single mutation for a particular antibody. The weaker results on these tasks reflect the increasing need to infer latent structural contacts and functional consequences that are not directly expressed in the input sequences.

\noindent\textbf{(3) Neither a single model nor reasoning capability yields a uniform advantage.} Gemini-3-Flash is strongest on region discovery and functional assessment, GPT-5.5 performs most consistently and leads epitope binning, while the nonreasoning Grok 4.20 leads epitope identification and escape assessment. This task-dependent ranking indicates that generic language-model reasoning does not by itself overcome the central bottlenecks of epitope understanding: long-range sequence grounding, residue-level correspondence, and antibody-specific structural inference. Overall, current LLMs provide partial competence on individual stages, but do not yet support the full epitope-centered antibody development workflow reliably.

\subsection{Antigen Length Sensitivity}

\begin{figure}[!htbp]
    \centering
    \includegraphics[width=\linewidth]{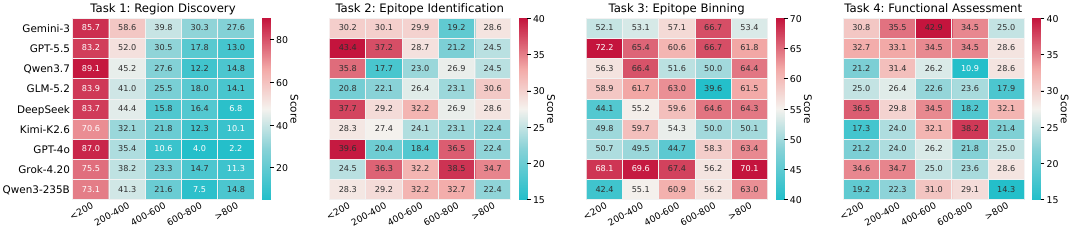}
    \caption{Performance of nine LLMs across antigen sequence length bins on EpiBench Tasks~1--4. Rows denote models and columns denote antigen length bins. Task~1 reports RegR@50, and Tasks~2--4 report accuracy. Each panel uses a task-specific color scale to emphasize within-task variation across sequence lengths.}
    \label{fig:length-bin-heatmap}
\end{figure}

Therapeutic antibody discovery involves antigens with diverse sequence scales, ranging from short domains to longer multi-domain proteins. This diversity raises a practical question: can LLMs understand epitope-related information from antigen sequences of different lengths, rather than only from compact inputs? To examine this question, we stratify Tasks~1--4 by antigen sequence length and report the results in Figure~\ref{fig:length-bin-heatmap}. Task~5 is not included because it evaluates mutation-level escape on a fixed antigen window instead of variable-length full-antigen inputs, so antigen length is not a meaningful axis for that task.

Task~1 exhibits the strongest dependence on antigen length. The average RegR@50 decreases from 81.3 for antigens shorter than 200 residues to 12.8 for antigens longer than 800 residues. This task asks the model to propose a small set of epitope residues before any antibody sequence is given, which is biologically close to identifying potentially targetable surface regions on an antigen. As the antigen becomes longer, the number of non-epitope residues, structural domains, and plausible surface patches increases, while the prediction budget remains fixed. The observed decline therefore reflects the increasing difficulty of recognizing sparse epitope-like signals against a larger antigen background. It also suggests that current LLMs have limited ability to preserve fine-grained residue position information when the relevant region occupies only a small fraction of a long sequence.

Tasks~2--4 provide a complementary view of how LLMs handle antigen sequences once the question supplies additional biological structure. Their performance changes are weaker and non-monotonic across length bins, and Task~3 does not decline on longer antigens. This suggests that LLMs can sometimes retain enough sequence-level information for candidate-based or comparative decisions even when the antigen is long. However, stable performance across length bins should not be interpreted as robust epitope understanding. In Task~2, the model must connect antibody sequence features with the correct candidate epitope; in Task~3, it must compare whether two antibodies are likely to target overlapping regions; and in Task~4, it must associate a candidate epitope with functional antibody evidence. These tasks require the model to organize sequence information around antibody specificity, epitope equivalence, and functional consequence. The length-stratified results therefore suggest that current LLMs are not limited only by raw input length; their broader challenge is to convert antigen and antibody sequences into biologically grounded epitope-level decisions.

\subsection{Effect of Explicit Reasoning}

Explicit reasoning prompts such as chain-of-thought (CoT) are commonly used to improve LLM performance, but their benefits are not guaranteed across tasks. Prior studies show that CoT effectiveness depends on task and prompt structure, that intermediate rationales may partly act as task specification rather than a universally useful reasoning mechanism, and that models vary in how strongly they condition on generated rationales \citep{madaan2023makes,lanham2023measuring}. Since the main EpiBench evaluation uses a unified zero-shot protocol that allows explicit intermediate reasoning, we compare it with an implicit-reasoning setting to quantify how much this reasoning format contributes when models interpret antigen and antibody sequences. In the implicit setting, models are asked to provide the final answer without generating intermediate reasoning steps.

\begin{table}[H]
    \centering
    \caption{Comparison of model performance between CoT prompting and the direct-answer setting. $\Delta=\text{Explicit thinking process}-\text{Implicit thinking process}$, and Avg. is the unweighted mean across tasks.}
    \label{tab:explicit-implicit}
    \vspace{0.5mm}
    \setlength{\tabcolsep}{2.4pt}
    \renewcommand{\arraystretch}{1.08}
    \resizebox{\textwidth}{!}{%
    \begin{tabular}{lcccccccccccc}
        \toprule
        &
        \multicolumn{2}{c}{\textbf{\makecell{T1: Region\\Discovery}}} &
        \multicolumn{2}{c}{\textbf{\makecell{T2: Epitope\\Identification}}} &
        \multicolumn{2}{c}{\textbf{\makecell{T3: Epitope\\Binning}}} &
        \multicolumn{2}{c}{\textbf{\makecell{T4: Functional\\Assessment}}} &
        \multicolumn{2}{c}{\textbf{\makecell{T5: Escape\\Risk}}} &
        \multicolumn{2}{c}{\textbf{Avg.}} \\
        \cmidrule(lr){2-3}\cmidrule(lr){4-5}\cmidrule(lr){6-7}\cmidrule(lr){8-9}\cmidrule(lr){10-11}\cmidrule(lr){12-13}
        \textbf{Model}
        & \textbf{Implicit} & \textbf{$\Delta$}
        & \textbf{Implicit} & \textbf{$\Delta$}
        & \textbf{Implicit} & \textbf{$\Delta$}
        & \textbf{Implicit} & \textbf{$\Delta$}
        & \textbf{Implicit} & \textbf{$\Delta$}
        & \textbf{Implicit} & \textbf{$\Delta$} \\
        \midrule
        Gemini-3-Flash & 48.9 & +8.0 & 29.7 & -1.4 & 62.7 & -9.2 & 31.2 & +4.4 & 47.8 & -1.6 & 44.0 & +0.0 \\
        GPT-5.5 & 43.5 & +6.2 & 28.2 & +3.7 & 60.0 & +7.3 & 28.5 & +4.7 & 46.9 & +5.3 & 41.4 & +5.4 \\
        Qwen3.7-Plus & 42.5 & +5.3 & 31.1 & -7.1 & 65.8 & -6.9 & 25.0 & +0.0 & 50.3 & -1.7 & 42.9 & -2.1 \\
        GLM-5.2 & 41.8 & +3.1 & 26.8 & -2.5 & 58.8 & -1.9 & 31.5 & -7.4 & 52.5 & -2.5 & 42.3 & -2.2 \\
        DeepSeek-V4-Pro & 42.0 & -10.2 & 25.7 & +5.1 & 51.9 & +3.1 & 26.2 & +4.1 & 55.6 & -7.2 & 40.3 & -1.0 \\
        Kimi-K2.6 & 33.4 & +3.2 & 27.1 & -1.7 & 53.8 & -1.9 & 25.3 & +1.8 & 34.4 & +6.9 & 34.8 & +1.7 \\
        GPT-4o & 21.0 & +17.5 & 28.2 & -2.8 & 59.2 & -8.1 & 26.5 & -2.6 & 53.6 & -1.4 & 37.7 & +0.5 \\
        Grok 4.20 & 41.3 & -0.6 & 29.1 & +4.5 & 60.8 & +5.0 & 30.0 & +0.0 & 50.0 & +4.7 & 42.2 & +2.7 \\
        Qwen3-235B & 35.2 & +5.3 & 29.1 & +0.3 & 60.8 & -11.5 & 22.6 & +1.8 & 53.6 & -6.9 & 40.3 & -2.2 \\
        \bottomrule
    \end{tabular}%
    }
\end{table}

Table~\ref{tab:explicit-implicit} shows that moving from implicit to explicit reasoning has model-dependent and task-dependent effects. GPT-5.5 is the most sensitive to this comparison, with explicit reasoning exceeding implicit reasoning by 5.4 points on average, whereas several other models show negligible or negative average differences. The clearest positive differences appear in Task~1, where most models perform better under the main explicit-reasoning protocol. This suggests that explicit reasoning can help organize the open-ended selection of sparse epitope-like regions from a full antigen sequence. However, the effect is much less consistent for Tasks~2--5. These tasks require the model to connect antibody sequences, candidate epitopes, functional annotations, or escape mutations to the underlying epitope relationship, and changing whether reasoning is expressed explicitly does not reliably change these sequence-grounded decisions. Overall, the comparison indicates that explicit reasoning can support some models and tasks, but EpiBench performance is not explained by prompting format alone; the central challenge remains biologically grounded sequence understanding.

\subsection{Failure Mode Analysis}

The preceding results suggest that current LLMs still have substantial room to improve in epitope-centered sequence understanding, but aggregate scores alone do not indicate which aspects of the reasoning process require improvement. We therefore analyze the reasoning traces of parseable but incorrect predictions under the main explicit-reasoning protocol. This analysis focuses on model behavior after a valid answer has been produced, separating reasoning-level failures from output-format failures.

\begin{figure}[H]
    \centering
    \includegraphics[width=0.92\linewidth]{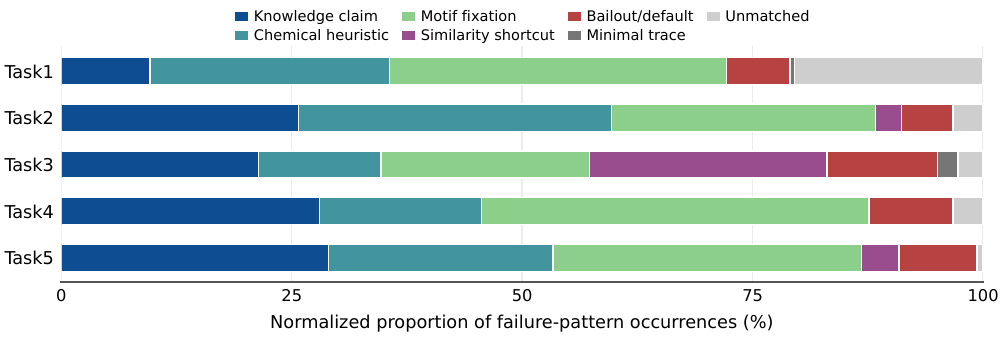}
    \caption{Failure-pattern composition across EpiBench tasks. We analyze parseable but incorrect CoT predictions from models with accessible reasoning traces and tag each trace with broad reasoning-pattern indicators. Because one trace may contain multiple patterns, each bar reports the normalized proportion of pattern occurrences within a task; unmatched denotes incorrect traces that do not trigger any predefined pattern.}
    \label{fig:reasoning-failure-modes}
\end{figure}

We use a rule-based, multi-label taxonomy to identify broad reasoning patterns that appear in incorrect traces. The categories include claims of memorized biological knowledge, generic physicochemical reasoning, fixation on sequence motifs or functional sites, CDR-sequence or germline-based shortcuts, and explicit fallback behavior such as guessing or choosing the safest option. Because one trace can contain several of these patterns, Figure~\ref{fig:reasoning-failure-modes} reports each task as a normalized distribution over pattern occurrences rather than mutually exclusive error classes.

The resulting patterns suggest that many incorrect answers are supported by plausible but weakly grounded biological rationales. Motif fixation is prominent in Tasks~1, 4, and 5, where models often emphasize recognizable sequence or functional-site cues even when these cues are insufficient to determine the correct epitope region, functional epitope, or escape effect. Knowledge claims and physicochemical heuristics are also frequent, especially in antibody-conditioned tasks, indicating that models often appeal to remembered antibody or antigen associations and generic residue-level interaction principles. Task~3 shows a distinct pattern: similarity shortcuts contribute a large fraction of failures, consistent with the task requiring epitope-bin equivalence rather than simple CDR-sequence or germline-based similarity. These observations do not imply that the detected patterns are the only causes of failure, but they show that incorrect LLM reasoning often relies on indirect biological proxies instead of consistently grounding predictions in antibody-specific epitope relationships.

\section{Discussion}

EpiBench occupies a distinct evaluation niche for biomedical LLMs: it asks whether general-purpose models can use antigen and antibody sequences to support epitope-centered decisions in antibody drug discovery. Unlike single-task epitope prediction datasets or broad protein benchmarks, EpiBench connects epitope localization, antibody-conditioned recognition, epitope binning, functional assessment, and escape assessment within a unified closed-book and automatically scorable framework. Our experiments show that current LLMs capture partial epitope-related signals, but still lag behind task-specific sequence models where such comparisons are available, show limited robustness across long antigen contexts, benefit inconsistently from explicit reasoning, and often rely on indirect biological proxies in incorrect reasoning traces. These findings position EpiBench not merely as a leaderboard, but as a diagnostic benchmark for identifying which aspects of antibody--antigen sequence understanding need improvement. By grounding evaluation in experimentally supported epitope evidence while preserving an LLM-ready formulation, EpiBench provides a practical testbed for developing and comparing sequence-aware biomedical LLMs, and for measuring progress toward reliable LLM-assisted antibody discovery.

\bibliographystyle{valhalla_references}
\bibliography{references}

@article{ahmed2026chimera,
  title={CHIMERA-bench: A benchmark dataset for epitope-specific antibody design},
  author={Ahmed, Mansoor and Taj, Nadeem and Khan, Imdad Ullah and Venkateswara, Hemanth and Patterson, Murray},
  journal={arXiv preprint arXiv:2603.13431},
  year={2026}
}

@article{haste2006prediction,
  title={Prediction of residues in discontinuous B-cell epitopes using protein 3D structures},
  author={Haste Andersen, Pernille and Nielsen, Morten and Lund, OLE},
  journal={Protein Science},
  volume={15},
  number={11},
  pages={2558--2567},
  year={2006},
  publisher={Wiley Online Library}
}

@article{ansari2010identification,
  title={Identification of conformational B-cell Epitopes in an antigen from its primary sequence},
  author={Ansari, Hifzur Rahman and Raghava, Gajendra PS},
  journal={Immunome research},
  volume={6},
  number={1},
  pages={6},
  year={2010},
  publisher={Springer}
}

@article{cao2022ba,
  title={BA. 2.12. 1, BA. 4 and BA. 5 escape antibodies elicited by Omicron infection},
  author={Cao, Yunlong and Yisimayi, Ayijiang and Jian, Fanchong and Song, Weiliang and Xiao, Tianhe and Wang, Lei and Du, Shuo and Wang, Jing and Li, Qianqian and Chen, Xiaosu and others},
  journal={Nature},
  volume={608},
  number={7923},
  pages={593--602},
  year={2022},
  publisher={Nature Publishing Group UK London}
}

@article{capel2022proteinglue,
  title={ProteinGLUE multi-task benchmark suite for self-supervised protein modeling},
  author={Capel, Henriette and Weiler, Robin and Dijkstra, Maurits and Vleugels, Reinier and Bloem, Peter and Feenstra, K Anton},
  journal={Scientific Reports},
  volume={12},
  number={1},
  pages={16047},
  year={2022},
  publisher={Nature Publishing Group UK London}
}

@article{chungyoun2024flab,
  title={FLAb: Benchmarking deep learning methods for antibody fitness prediction},
  author={Chungyoun, Michael and Ruffolo, Jeffrey and Gray, Jeffrey},
  journal={BioRxiv},
  pages={2024--01},
  year={2024},
  publisher={Cold Spring Harbor Laboratory}
}

@article{clifford2022bepipred,
  title={BepiPred-3.0: Improved B-cell epitope prediction using protein language models},
  author={Clifford, Joakim N{\o}ddeskov and H{\o}ie, Magnus Haraldson and Deleuran, Sebastian and Peters, Bjoern and Nielsen, Morten and Marcatili, Paolo},
  journal={Protein Science},
  volume={31},
  number={12},
  pages={e4497},
  year={2022},
  publisher={Wiley Online Library}
}

@article{dallago2021flip,
  title={FLIP: Benchmark tasks in fitness landscape inference for proteins},
  author={Dallago, Christian and Mou, Jody and Johnston, Kadina E and Wittmann, Bruce J and Bhattacharya, Nicholas and Goldman, Samuel and Madani, Ali and Yang, Kevin K},
  journal={bioRxiv},
  pages={2021--11},
  year={2021},
  publisher={Cold Spring Harbor Laboratory}
}

@inproceedings{didi2026flip2,
  title={FLIP2: Expanding protein fitness landscape benchmarks for real-world machine learning applications},
  author={Didi, Kieran and Alamdari, Sarah and Lu, Alex Xijie and Wittmann, Bruce James and Johnston, Kadina E and Amini, Ava P and Madani, Ali and Czeneszew, Maya and Dallago, Christian and Yang, Kevin K},
  booktitle={Forty-third International Conference on Machine Learning},
  year={2026}
}

@article{dunbar2014sabdab,
  title={SAbDab: the structural antibody database},
  author={Dunbar, James and Krawczyk, Konrad and Leem, Jinwoo and Baker, Terry and Fuchs, Angelika and Georges, Guy and Shi, Jiye and Deane, Charlotte M},
  journal={Nucleic acids research},
  volume={42},
  number={D1},
  pages={D1140--D1146},
  year={2014},
  publisher={Oxford University Press}
}

@article{el2008predicting,
  title={Predicting linear B-cell epitopes using string kernels},
  author={EL-Manzalawy, Yasser and Dobbs, Drena and Honavar, Vasant},
  journal={Journal of Molecular Recognition: An Interdisciplinary Journal},
  volume={21},
  number={4},
  pages={243--255},
  year={2008},
  publisher={Wiley Online Library}
}

@article{elnaggar2021others,
  title={others Prottrans: Toward understanding the language of life through self-supervised learning},
  author={Elnaggar, Ahmed and Heinzinger, Michael and Dallago, Christian and Rehawi, Ghalia and Wang, Yu and Jones, Llion and Gibbs, Tom and Feher, Tamas and Angerer, Christoph and Steinegger, Martin and others},
  journal={IEEE transactions on pattern analysis and machine intelligence},
  volume={44},
  number={10},
  pages={7112--7127},
  year={2021}
}

@article{emini1985induction,
  title={Induction of hepatitis A virus-neutralizing antibody by a virus-specific synthetic peptide},
  author={Emini, Emilio A and Hughes, JOSEPH V and Perlow, DoS and Boger, J},
  journal={Journal of virology},
  volume={55},
  number={3},
  pages={836--839},
  year={1985}
}

@inproceedings{fang2024mol,
  title={Mol-instructions: A large-scale biomolecular instruction dataset for large language models},
  author={Fang, Yin and Liang, Xiaozhuan and Zhang, Ningyu and Liu, Kangwei and Huang, Rui and Chen, Zhuo and Fan, Xiaohui and Chen, Huajun},
  booktitle={International Conference on Learning Representations},
  volume={2024},
  pages={48221--48251},
  year={2024}
}

@article{huang2022artificial,
  title={Artificial intelligence foundation for therapeutic science},
  author={Huang, Kexin and Fu, Tianfan and Gao, Wenhao and Zhao, Yue and Roohani, Yusuf and Leskovec, Jure and Coley, Connor W and Xiao, Cao and Sun, Jimeng and Zitnik, Marinka},
  journal={Nature chemical biology},
  volume={18},
  number={10},
  pages={1033--1036},
  year={2022},
  publisher={Nature Publishing Group US New York}
}

@article{kolaskar1990semi,
  title={A semi-empirical method for prediction of antigenic determinants on protein antigens},
  author={Kolaskar, Ashok S and Tongaonkar, Prasad C},
  journal={FEBS letters},
  volume={276},
  number={1-2},
  pages={172--174},
  year={1990},
  publisher={Elsevier}
}

@article{lanham2023measuring,
  title={Measuring faithfulness in chain-of-thought reasoning},
  author={Lanham, Tamera and Chen, Anna and Radhakrishnan, Ansh and Steiner, Benoit and Denison, Carson and Hernandez, Danny and Li, Dustin and Durmus, Esin and Hubinger, Evan and Kernion, Jackson and others},
  journal={arXiv preprint arXiv:2307.13702},
  year={2023}
}

@article{larsen2006improved,
  title={Improved method for predicting linear B-cell epitopes},
  author={Larsen, Jens Erik Pontoppidan and Lund, Ole and Nielsen, Morten},
  journal={Immunome research},
  volume={2},
  number={1},
  pages={2},
  year={2006},
  publisher={Springer}
}

@article{lin2023evolutionary,
  title={Evolutionary-scale prediction of atomic-level protein structure with a language model},
  author={Lin, Zeming and Akin, Halil and Rao, Roshan and Hie, Brian and Zhu, Zhongkai and Lu, Wenting and Smetanin, Nikita and Verkuil, Robert and Kabeli, Ori and Shmueli, Yaniv and others},
  journal={Science},
  volume={379},
  number={6637},
  pages={1123--1130},
  year={2023},
  publisher={American Association for the Advancement of Science}
}

@article{liu2024asep,
  title={AsEP: Benchmarking deep learning methods for antibody-specific epitope prediction},
  author={Liu, Chunan and Denzler, Lilian and Chen, Yihong and Martin, Andrew and Paige, Brooks},
  journal={Advances in Neural Information Processing Systems},
  volume={37},
  pages={11700--11734},
  year={2024}
}

@inproceedings{madaan2023makes,
  title={What makes chain-of-thought prompting effective? a counterfactual study},
  author={Madaan, Aman and Hermann, Katherine and Yazdanbakhsh, Amir},
  booktitle={Findings of the Association for Computational Linguistics: EMNLP 2023},
  pages={1448--1535},
  year={2023}
}

@article{madani2023large,
  title={Large language models generate functional protein sequences across diverse families},
  author={Madani, Ali and Krause, Ben and Greene, Eric R and Subramanian, Subu and Mohr, Benjamin P and Holton, James M and Olmos Jr, Jose Luis and Xiong, Caiming and Sun, Zachary Z and Socher, Richard and others},
  journal={Nature biotechnology},
  volume={41},
  number={8},
  pages={1099--1106},
  year={2023},
  publisher={Nature Publishing Group US New York}
}

@article{mollon2025exploring,
  title={Exploring Large Protein Language Models in Constrained Evaluation Scenarios within the FLIP Benchmark},
  author={Mollon, Manuel F and Gonzalez-Rodriguez, Joaquin and Lozano-Diez, Alicia and Ramos, Daniel and Toledano, Doroteo T},
  journal={arXiv preprint arXiv:2501.18223},
  year={2025}
}

@article{nijkamp2023progen2,
  title={Progen2: exploring the boundaries of protein language models},
  author={Nijkamp, Erik and Ruffolo, Jeffrey A and Weinstein, Eli N and Naik, Nikhil and Madani, Ali},
  journal={Cell systems},
  volume={14},
  number={11},
  pages={968--978},
  year={2023},
  publisher={Elsevier}
}

@inproceedings{notin2022tranception,
  title={Tranception: protein fitness prediction with autoregressive transformers and inference-time retrieval},
  author={Notin, Pascal and Dias, Mafalda and Frazer, Jonathan and Marchena-Hurtado, Javier and Gomez, Aidan N and Marks, Debora and Gal, Yarin},
  booktitle={International Conference on Machine Learning},
  pages={16990--17017},
  year={2022},
  organization={PMLR}
}

@article{notin2023proteingym,
  title={Proteingym: Large-scale benchmarks for protein fitness prediction and design},
  author={Notin, Pascal and Kollasch, Aaron and Ritter, Daniel and Van Niekerk, Lood and Paul, Steffanie and Spinner, Han and Rollins, Nathan and Shaw, Ada and Orenbuch, Rose and Weitzman, Ruben and others},
  journal={Advances in neural information processing systems},
  volume={36},
  pages={64331--64379},
  year={2023}
}

@article{ponomarenko2008ellipro,
  title={ElliPro: a new structure-based tool for the prediction of antibody epitopes},
  author={Ponomarenko, Julia and Bui, Huynh-Hoa and Li, Wei and Fusseder, Nicholas and Bourne, Philip E and Sette, Alessandro and Peters, Bjoern},
  journal={BMC bioinformatics},
  volume={9},
  number={1},
  pages={514},
  year={2008},
  publisher={Springer}
}

@article{rao2019evaluating,
  title={Evaluating protein transfer learning with TAPE},
  author={Rao, Roshan and Bhattacharya, Nicholas and Thomas, Neil and Duan, Yan and Chen, Xi and Canny, John and Abbeel, Pieter and Song, Yun S},
  journal={Biorxiv},
  pages={676825},
  year={2019},
  publisher={Cold Spring Harbor Laboratory}
}

@article{rives2021biological,
  title={Biological structure and function emerge from scaling unsupervised learning to 250 million protein sequences},
  author={Rives, Alexander and Meier, Joshua and Sercu, Tom and Goyal, Siddharth and Lin, Zeming and Liu, Jason and Guo, Demi and Ott, Myle and Zitnick, C Lawrence and Ma, Jerry and others},
  journal={Proceedings of the national academy of sciences},
  volume={118},
  number={15},
  pages={e2016239118},
  year={2021},
  publisher={National Academy of Sciences}
}

@article{saha2006prediction,
  title={Prediction of continuous B-cell epitopes in an antigen using recurrent neural network},
  author={Saha, Sudipto and Raghava, Gajendra Pal Singh},
  journal={Proteins: Structure, Function, and Bioinformatics},
  volume={65},
  number={1},
  pages={40--48},
  year={2006},
  publisher={Wiley Online Library}
}

@article{singhal2022large,
  title={Large language models encode clinical knowledge},
  author={Singhal, Karan and Azizi, Shekoofeh and Tu, Tao and Mahdavi, S Sara and Wei, Jason and Chung, Hyung Won and Scales, Nathan and Tanwani, Ajay and Cole-Lewis, Heather and Pfohl, Stephen and others},
  journal={arXiv preprint arXiv:2212.13138},
  year={2022}
}

@article{steinegger2017mmseqs2,
  title={MMseqs2 enables sensitive protein sequence searching for the analysis of massive data sets},
  author={Steinegger, Martin and S{\"o}ding, Johannes},
  journal={Nature biotechnology},
  volume={35},
  number={11},
  pages={1026--1028},
  year={2017},
  publisher={Nature Publishing Group US New York}
}

@article{sun2009seppa,
  title={SEPPA: a computational server for spatial epitope prediction of protein antigens},
  author={Sun, Jing and Wu, Di and Xu, Tianlei and Wang, Xiaojing and Xu, Xiaolian and Tao, Lin and Li, YX and Cao, Zhi-Wei},
  journal={Nucleic acids research},
  volume={37},
  number={suppl\_2},
  pages={W612--W616},
  year={2009},
  publisher={Oxford University Press}
}

@article{tan2024peta,
  title={PETA: evaluating the impact of protein transfer learning with sub-word tokenization on downstream applications},
  author={Tan, Yang and Li, Mingchen and Zhou, Ziyi and Tan, Pan and Yu, Huiqun and Fan, Guisheng and Hong, Liang},
  journal={Journal of Cheminformatics},
  volume={16},
  number={1},
  pages={92},
  year={2024},
  publisher={Springer}
}

@inproceedings{townshend2021atom3d,
  title={Atom3d: Tasks on molecules in three dimensions},
  author={Townshend, Raphael John Lamarre and V{\"o}gele, Martin and Suriana, Patricia Adriana and Derry, Alexander and Powers, Alexander and Laloudakis, Yianni and Balachandar, Sidhika and Jing, Bowen and Anderson, Brandon M and Eismann, Stephan and others},
  booktitle={Thirty-fifth Conference on Neural Information Processing Systems Datasets and Benchmarks Track (Round 1)},
  year={2021}
}

@article{van20262025,
  title={2025 ginkgo datapoints antibody developability competition outcomes: limited model performance and a call for data standardization},
  author={van Niekerk, Lood and Moller, Joshua and Ritter, Seth and Quintero-Cadena, Porfirio and Cohen, Rich and Channing, Georgia and Chungyoun, Michael and Rand, Laura and Smith, Alexander and Bhatt, Aanal and others},
  journal={MAbs},
  volume={18},
  number={1},
  pages={2634216},
  year={2026},
  publisher={Taylor \& Francis}
}

@article{vita2025immune,
  title={The immune epitope database (IEDB): 2024 update},
  author={Vita, Randi and Blazeska, Nina and Marrama, Daniel and IEDB Curation Team Members Shackelford Deborah Zalman Leora Foos Gabriele Zarebski Laura Chan Kenneth Reardon Brian Fitzpatrick Sidne Busse Matthew Coleman Sara Sedwick Caitlin Edwards Lindy MacFarlane Catriona Ennis Marcus and Duesing, Sebastian and Bennett, Jason and Greenbaum, Jason and De Almeida Mendes, Marcus and Mahita, Jarjapu and Wheeler, Daniel K and others},
  journal={Nucleic Acids Research},
  volume={53},
  number={D1},
  pages={D436--D443},
  year={2025},
  publisher={Oxford University Press}
}

@article{waury2025comparison,
  title={Comparison of sequence-and structure-based antibody clustering approaches on simulated repertoire sequencing data},
  author={Waury, Katharina and Lelieveld, Stefan and Abeln, Sanne and van den Ham, Henk-Jan},
  journal={PLoS Computational Biology},
  volume={21},
  number={5},
  pages={e1013057},
  year={2025},
  publisher={Public Library of Science San Francisco, CA USA}
}

@article{xu2022peer,
  title={Peer: a comprehensive and multi-task benchmark for protein sequence understanding},
  author={Xu, Minghao and Zhang, Zuobai and Lu, Jiarui and Zhu, Zhaocheng and Zhang, Yangtian and Chang, Ma and Liu, Runcheng and Tang, Jian},
  journal={Advances in Neural Information Processing Systems},
  volume={35},
  pages={35156--35173},
  year={2022}
}

@article{zhao2025benchmark,
  title={Benchmark for antibody binding affinity maturation and design},
  author={Zhao, Xinyan and Tang, Yi-Ching and Singh, Akshita and Cantu, Victor J and An, KwanHo and Lee, Junseok and Stogsdill, Adam E and Ramesh, Ashwin Kumar and An, Zhiqiang and Jiang, Xiaoqian and others},
  journal={arXiv e-prints},
  pages={arXiv--2506},
  year={2025}
}

@inproceedings{zhuo2024protllm,
  title={Protllm: An interleaved protein-language llm with protein-as-word pre-training},
  author={Zhuo, Le and Chi, Zewen and Xu, Minghao and Huang, He-Yan and Zhao, Jianan and Zheng, Heqi and He, Conghui and Mao, Xian-Ling and Zhang, Wentao},
  booktitle={Proceedings of the 62nd Annual Meeting of the Association for Computational Linguistics (Volume 1: Long Papers)},
  pages={8950--8963},
  year={2024}
}

@article{wang2023unirna,
  title={UNI-RNA: universal pre-trained models revolutionize RNA research},
  author={Wang, Xi and Gu, Ruichu and Chen, Zhiyuan and Li, Yongge and Ji, Xiaohong and Ke, Guolin and Wen, Han},
  journal={bioRxiv},
  pages={2023--07},
  year={2023},
  publisher={Cold Spring Harbor Laboratory}
}

@inproceedings{wu2024fafe,
  title={FAFE: Immune Complex Modeling with Geodesic Distance Loss on Noisy Group Frames},
  author={Wu, Ruidong and Guo, Ruihan and Wang, Rui and Luo, Shitong and Xu, Yue and Li, Jiahan and Ma, Jianzhu and Liu, Qiang and Luo, Yunan and Peng, Jian},
  booktitle={Proceedings of the 41st International Conference on Machine Learning},
  volume={235},
  pages={53422--53442},
  year={2024},
  publisher={PMLR}
}

@article{xu2026sakepp,
  title={SAKE-PP: A Spatial-Attention Equivariant Network for Accurate Ranking of Protein-Protein Interaction Models},
  author={Xu, Yuzhi and Xia, Wei and Zhang, Chao and Liu, Xinxin and Ju, Cheng-Wei and Dai, Xuhang and Xie, Pujun and Wang, Yuanqing and Chen, Guangyong and Zhang, John Z. H.},
  journal={JACS Au},
  volume={6},
  number={5},
  pages={2846--2856},
  year={2026},
  publisher={American Chemical Society (ACS)}
}

@inproceedings{li2025pephar,
  title={Hotspot-Driven Peptide Design via Multi-Fragment Autoregressive Extension},
  author={Li, Jiahan and Chen, Tong and Luo, Shitong and Cheng, Chaoran and Guan, Jiaqi and Guo, Ruihan and Wang, Sheng and Liu, Ge and Peng, Jian and Ma, Jianzhu},
  booktitle={The Thirteenth International Conference on Learning Representations},
  year={2025}
}

@inproceedings{guo2024retrieval,
  title={Enhancing Protein Mutation Effect Prediction through a Retrieval-Augmented Framework},
  author={Guo, Ruihan and Wang, Rui and Wu, Ruidong and Ren, Zhizhou and Li, Jiahan and Luo, Shitong and Wu, Zuofan and Liu, Qiang and Peng, Jian and Ma, Jianzhu},
  booktitle={Advances in Neural Information Processing Systems},
  volume={37},
  pages={49130--49153},
  year={2024}
}

\clearpage
\appendix
\EpibenchAppendixHeadingSpacing
\EpibenchAppendixHeadingSpacing

\section{Biological Concepts and Task Formalization}

\subsection{Epitope-related Biological Concepts}

EpiBench evaluates sequence based reasoning about antigen--antibody recognition, so we first clarify the biological concepts used throughout the benchmark. An antigen is the molecular target recognized by an antibody, and an antibody binds an antigen through its variable domains, especially the complementarity determining regions (CDRs). The antigen residues recognized by an antibody are called the epitope, while the antibody residues involved in recognition are often referred to as the paratope. In structural settings, an epitope can be represented as a set of antigen residues that are spatially close to antibody CDR residues; in sequence based assay settings, it can also appear as a linear peptide segment with experimentally measured binding or functional evidence.

Epitopes are central to several downstream decisions in antibody discovery. A targetable antigen region denotes an antigen region repeatedly or plausibly engaged by antibodies and is useful for early antigen characterization. Antibody conditioned epitope identification asks which region is recognized by a particular antibody, rather than which region is generally targetable. Epitope binning groups antibodies according to whether they bind the same or strongly overlapping antigen regions, which supports panel selection and mechanism comparison. A functional epitope is an epitope associated with a downstream biological effect, such as inhibition, protection, or cytotoxicity, rather than binding evidence alone. Finally, an escape mutation is an antigen mutation that reduces or disrupts recognition by a given antibody. These concepts motivate the five tasks in EpiBench.

\subsection{Formal Definition of EpiBench Tasks}

EpiBench converts the above biological concepts into five automatically scorable sequence based tasks. Each task removes direct identifiers and provides only the sequence level evidence needed for the corresponding decision. Table~\ref{tab:appendix-task-formalization} summarizes the biological question, model input, and expected output for each task.

\begin{table}[H]
\centering
\footnotesize
\setlength{\tabcolsep}{3.5pt}
\renewcommand{\arraystretch}{1.15}
\caption{Formal definition of the five EpiBench tasks. Each task represents one epitope centered decision in antibody drug discovery and is formulated with sequence based inputs and automatically scorable outputs.}
\label{tab:appendix-task-formalization}
\begin{tabular}{p{0.10\linewidth}p{0.32\linewidth}p{0.28\linewidth}p{0.22\linewidth}}
\toprule
\textbf{Task} & \textbf{Biological question} & \textbf{Input} & \textbf{Output} \\
\midrule
Task1 & Which antigen residues form targetable regions? & Antigen sequence & Epitope residue indices \\
Task2 & Which candidate epitope is bound by a given antibody? & Antigen sequence, antibody sequences, and candidate epitope regions & One candidate epitope \\
Task3 & Do two antibodies belong to the same epitope bin? & Antigen sequence and two antibody sequence pairs & Same or different bin \\
Task4 & Which candidate epitope is associated with the specified functional effect? & Antigen sequence, candidate epitope regions, and functional context & One functional epitope \\
Task5 & Does an antigen mutation cause escape from a given antibody? & Antibody sequences, wild type antigen region, and mutation & Escape or non escape \\
\bottomrule
\end{tabular}
\end{table}

The tasks differ in output format because they correspond to different experimental decisions. Task1 requires residue level localization over an antigen sequence. Tasks 2 and 4 are formulated as multiple choice decisions because the biological setting often involves selecting a binding or functional region from candidate epitopes. Tasks 3 and 5 are formulated as binary judgments because epitope binning and escape assessment naturally ask whether a relation holds for a specific antibody pair or antibody--mutation pair. This formulation keeps the benchmark close to practical antibody discovery questions while allowing deterministic scoring.

\section{Dataset Construction and Quality Control}

\subsection{Source Datasets}

EpiBench is constructed from experimentally grounded resources that provide complementary evidence for epitope centered antibody discovery tasks. Structural antibody--antigen complexes provide residue level contact labels for Tasks 1 to 3, curated B cell assay records provide functional epitope evidence for Task4, and deep mutational scanning measurements provide mutation specific escape labels for Task5. Table~\ref{tab:source-datasets} summarizes the source datasets and their roles in EpiBench.

\begin{table}[H]
\centering
\footnotesize
\setlength{\tabcolsep}{4pt}
\renewcommand{\arraystretch}{1.12}
\caption{Source datasets used to construct EpiBench. Each source contributes a distinct experimental evidence type for one or more benchmark tasks.}
\label{tab:source-datasets}
\begin{tabular}{p{0.18\linewidth}p{0.31\linewidth}p{0.15\linewidth}p{0.28\linewidth}}
\toprule
\textbf{Source} & \textbf{Evidence type} & \textbf{Tasks} & \textbf{Role in EpiBench} \\
\midrule
AsEP~\citep{liu2024asep} & Curated antibody--antigen structural contacts & Tasks 1--3 & Provides antibody-specific epitope footprints and contact based residue labels \\
SAbDab~\citep{dunbar2014sabdab} & Antibody--antigen complex structures & Tasks 1--3 & Expands the structural record pool beyond curated AsEP examples \\
IEDB~\citep{vita2025immune} & Curated B cell epitope assay records & Task4 & Provides functional and binding-only epitope evidence for same-antigen candidate construction \\
UniProt-linked antigen sequences & Parent protein sequences for linear peptides & Task4 & Maps IEDB peptide epitopes back to antigen sequence coordinates \\
DMS escape measurements~\citep{cao2022ba} & Antibody-specific mutation escape scores & Task5 & Provides escape and non-escape labels for antibody--mutation pairs \\
\bottomrule
\end{tabular}
\end{table}

\subsection{Task-specific Construction Procedures}

\subsubsection{Task1: Targetable Region Discovery}

Task1 is built from the shared structural record pool used by Tasks 1 to 3, containing 3346 antibody antigen complexes from curated AsEP records and newly processed SAbDab structures \citep{liu2024asep,dunbar2014sabdab}. Following the contact based definition used in AsEP, an epitope residue is defined as an antigen surface residue whose heavy atoms are within 4.5~\AA{} of antibody CDR heavy atoms \citep{liu2024asep}. We retain standard protein antigens with valid mapped epitopes, cluster antigen sequences with MMseqs2 at 95\% identity and 0.8 coverage \citep{steinegger2017mmseqs2}, merge domain and full length variants using a 0.9 identity and 0.8 coverage search criterion, and cluster antibodies at 95\% identity using concatenated heavy and light chain variable domains. For each antigen group, the longest antigen is used as the reference, one representative complex is selected for each distinct antibody, and all representative footprints are mapped to the reference sequence. The Task1 label is the union of mapped footprints, with patch labels obtained by grouping overlapping footprints, while antigen groups with fewer than two distinct antibodies or excessive surface coverage are removed. The final set contains 295 antigens, and Algorithm~\ref{alg:task1-construction} summarizes the shared structural construction procedure.

\begin{center}
\footnotesize
\begingroup
\setlength{\tabcolsep}{0pt}
\setlength{\aboverulesep}{0pt}
\setlength{\belowrulesep}{0.25ex}
\renewcommand{\arraystretch}{1.15}
\begin{tabular}{p{0.95\linewidth}}
\toprule
\EpiAlgorithmCaption{Targetable Region Discovery Dataset Construction}{alg:task1-construction} \\
\midrule
\begin{minipage}{0.95\linewidth}
\textbf{Require:} Structural records $\mathcal{R}$, where each record $r$ contains antigen $A_r$, antibody $B_r=H_r\Vert L_r$, epitope $E_r$, and antigen surface $S_r$.\\
\textbf{Ensure:} Task1 set $\mathcal{D}_{1}=\{(A_{\mathrm{ref}},U,P)\}$ with union epitope $U$ and patch labels $P$.\\[0.2em]
\begin{tabular}{@{}r@{\hspace{0.6em}}p{0.88\linewidth}@{}}
1: & $\mathcal{R}' \leftarrow \operatorname{Filter}(\mathcal{R}; |A_r|\geq50, E_r\neq\emptyset)$.\\
2: & $\mathcal{G} \leftarrow \operatorname{ClusterAntigens}(\mathcal{R}'; (0.95,0.8), (0.9,0.8))$.\\
3: & $\mathcal{D}_{1} \leftarrow \emptyset$.\\
4: & \textbf{for} antigen group $G\in\mathcal{G}$ \textbf{do}\\
5: & \quad $A_{\mathrm{ref}} \leftarrow \operatorname{LongestAntigen}(G)$.\\
6: & \quad $\mathcal{C}_{\mathrm{ab}} \leftarrow \operatorname{ClusterAntibodies}(G; B_r, 0.95)$.\\
7: & \quad \textbf{if} $|\mathcal{C}_{\mathrm{ab}}|<2$ \textbf{then continue}.\\
8: & \quad $Q \leftarrow \{\operatorname{BestFootprint}(C):C\in\mathcal{C}_{\mathrm{ab}}\}$.\\
9: & \quad $(\widetilde{E}_r,\widetilde{S}_r) \leftarrow \operatorname{MapToRef}(E_r,S_r,A_{\mathrm{ref}})$ for $r\in Q$.\\
10: & \quad $U \leftarrow \bigcup_{r\in Q}\widetilde{E}_r$, $S \leftarrow \bigcup_{r\in Q}\widetilde{S}_r$.\\
11: & \quad \textbf{if} $|U|/|S|>0.5$ \textbf{then continue}.\\
12: & \quad $P \leftarrow \operatorname{BuildPatches}(\{\widetilde{E}_r:r\in Q\}; \operatorname{OvCoeff}\geq0.5)$.\\
13: & \quad $\mathcal{D}_{1}\leftarrow\mathcal{D}_{1}\cup\{(A_{\mathrm{ref}},U,P)\}$.\\
14: & \textbf{end for}\\
15: & \textbf{return} $\mathcal{D}_{1}$\\
\end{tabular}
\end{minipage}\\
\bottomrule
\end{tabular}
\endgroup
\end{center}

\subsubsection{Task2: Antibody-conditioned Epitope Identification}

Task2 reuses the structural records and reference mapped antibody footprints from Task1, but keeps each representative antibody footprint as an antibody specific label rather than collapsing footprints into an antigen level union. Footprints with fewer than three residues are removed, and overlapping footprints on the same antigen are grouped into candidate regions using an overlap coefficient threshold of 0.5. We construct multiple choice items in two tiers: Tier A uses antigen groups with at least four real candidate regions and samples distractors from other real regions on the same antigen, while Tier B uses the largest valid footprint as the answer and samples non overlapping synthetic distractors matched by size, span, and segmented shape. Candidate orders are shuffled and answer letters are balanced, yielding 354 examples with 254 real distractor items and 100 synthetic distractor items.

\begin{table}[H]
\centering
\footnotesize
\setlength{\tabcolsep}{4pt}
\renewcommand{\arraystretch}{1.12}
\caption{De-identification and shortcut-control strategies used during EpiBench construction.}
\label{tab:leakage-control}
\begin{tabular}{p{0.38\linewidth}p{0.17\linewidth}p{0.36\linewidth}}
\toprule
\textbf{Control strategy} & \textbf{Applied tasks} & \textbf{Purpose} \\
\midrule
Remove PDB IDs, antibody names, antigen names, accession numbers, variant names, and publication metadata & Tasks 1--5 & Reduce direct lookup from prompt identifiers or source-record metadata \\
Use sequence-level prompts with position-numbered antigens, antibody VH/VL sequences, CDRs, candidate residues, or mutations as needed & Tasks 1--5 & Keep only the evidence required for the target biological decision \\
Cluster antigen sequences and merge near-duplicate antigen records during structural task construction & Tasks 1--3 & Reduce redundancy from closely related structural records \\
Cluster antibodies or remove near-duplicate antibody pairs using variable-domain or CDR similarity & Tasks 1--3 & Reduce repeated antibody-derived shortcuts \\
Randomize and balance candidate option orders & Tasks 2, 4 & Reduce answer-position bias in multiple-choice questions \\
Match same-bin and different-bin antibody pairs within CDR-similarity strata & Task3 & Reduce shortcuts based only on antibody sequence similarity \\
Match escape and non-escape examples at the mutation level & Task5 & Reduce shortcuts based only on mutation identity \\
Balance binary labels and multiple-choice answer letters where applicable & Tasks 2--5 & Reduce class-prior and option-prior biases \\
\bottomrule
\end{tabular}
\end{table}

\subsubsection{Task3: Epitope Binning}

Task3 is constructed by grouping structural records with exactly identical antigen sequences and comparing antibody epitope residue sets in the same coordinate system. For each antibody pair, we compute the epitope Jaccard index $J=|E_a\cap E_b|/|E_a\cup E_b|$ and label pairs as same bin when $J\geq0.75$ following the structure based binning criterion in \citet{waury2025comparison}; pairs with no shared residues are labeled as different bin, and partially overlapping pairs are discarded. Near duplicate antibodies are removed by CDR sequence similarity, and candidate pairs are sampled within CDR similarity quantile bins so that same and different pairs have matched antibody similarity profiles. The released set contains 260 balanced pairs, with 130 same bin and 130 different bin examples, and is stratified by antigen length tertiles.

\subsubsection{Task4: Functional Epitope Assessment}

Task4 is constructed from IEDB B cell assay records with linear peptide epitopes and UniProt linked parent antigens \citep{vita2025immune}. We retain peptides of length 6 to 40 that contain only standard amino acids, occur uniquely in the antigen sequence, and belong to antigens no longer than 1200 residues. Assay names are mapped into functional mechanisms, including inhibition, protection, and cytotoxicity, and each epitope is assigned its strongest evidence type among functional positive, functional negative, and binding only. Each multiple choice item uses a functional positive epitope as the answer and selects distractors from same antigen functional negative or binding only epitopes when available, otherwise using non overlapping synthetic windows matched by length and sequence properties. Antigens are clustered to limit redundancy, answer letters are balanced, and the final set contains 340 items with mechanism categories inhibition, protection, and cytotoxicity.

\subsubsection{Task5: Escape Assessment}

Task5 is built from deep mutational scanning escape measurements for SARS CoV 2 RBD antibodies \citep{cao2022ba}. For each antibody, raw escape scores are normalized by the maximum escape value observed for that antibody, and mutations are labeled as escape when the normalized score is at least 0.50 and non escape when it is at most 0.05, with intermediate cases discarded. To reduce shortcuts from mutation identity, we retain only mutations that have both escape and non escape antibody examples and sample one example from each class for the same mutation. The antigen input is rendered as the RBD window with local one based numbering, while native identifiers, antibody names, antigen names, and variant names are removed from prompts. The final set contains 360 balanced examples across five wild type residue classes, with 180 escape and 180 non escape labels.

\begin{figure}[H]
    \centering
    \includegraphics[width=\linewidth]{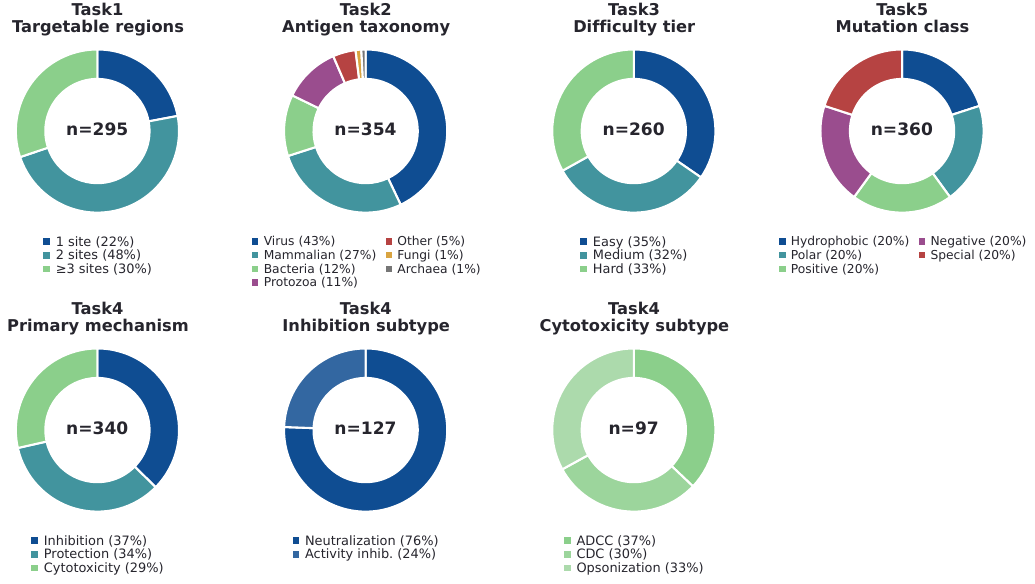}
    \caption{Fine grained subcategory distributions in EpiBench. Each task is assigned a primary subcategory for stratified evaluation. Task4 also includes secondary mechanism distributions for inhibition and cytotoxicity, while protection is retained at the primary mechanism level.}
    \label{fig:subcategory-pies}
\end{figure}

\subsection{De-identification and Leakage Control}

EpiBench is designed as a closed book, sequence based benchmark, so the construction process removes direct identifiers and reduces shortcut signals that could make the tasks solvable without epitope reasoning. We do not claim that any benchmark can fully eliminate pretraining memorization of public biological sequences. Instead, our controls target prompt level leakage and distributional shortcuts: the rendered prompts exclude record identifiers and metadata, while task specific sampling reduces answer position, class prior, antibody similarity, and mutation identity biases. Table~\ref{tab:leakage-control} summarizes these controls.

\subsection{Dataset Statistics and Subcategory Definitions}

To support stratified evaluation, each EpiBench task is assigned one primary subcategory label along an objective and task relevant axis. Task1 is stratified by the number of targetable antigen regions, computed from overlap based epitope patches. Task2 is stratified by antigen taxonomic category, which captures source organism diversity. Task3 is stratified by difficulty tier using antigen sequence length tertiles, reflecting the larger search space induced by longer antigens. Task4 is stratified by the primary functional mechanism of the correct epitope, including inhibition, protection, and cytotoxicity. Task5 is stratified by the physicochemical class of the wild type residue being mutated. These subcategory labels are used for reporting model behavior across biological and difficulty conditions, and are not included in the model prompts or answer labels.

Our construction increases subcategory diversity where possible, but does not force strict uniformity across all tasks because some categories are rare after experimental evidence filtering and quality control. Task5 is intentionally balanced across the five mutation residue classes, whereas Tasks 1 to 4 retain the natural distribution of high confidence samples after filtering. This design preserves data reliability while enabling post hoc analysis of whether model performance is concentrated in majority categories. For Task4, we additionally record secondary mechanisms for inhibition and cytotoxicity, while protection is kept at the primary mechanism level because no reliable secondary mechanism partition is used in this benchmark. The resulting fine grained distributions are shown in Figure~\ref{fig:subcategory-pies}.

\subsection{Answer Distribution and Choice Randomization}

\begin{figure}[H]
    \centering
    \includegraphics[width=\linewidth]{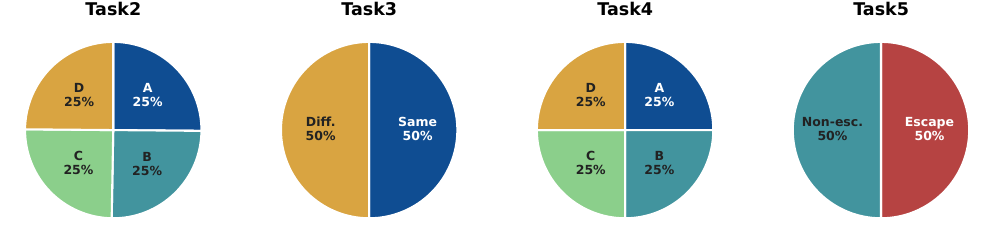}
    \caption{Answer distributions for EpiBench tasks with discrete outputs. Tasks 2 and 4 use four-way multiple-choice answers, while Tasks 3 and 5 use binary labels. Task1 is excluded because it is a residue-level prediction task without fixed answer options.}
    \label{fig:answer-distribution}
\end{figure}

For tasks with discrete answer formats, we balance answer distributions to reduce position and class-prior shortcuts. Tasks 2 and 4 randomize candidate order and maintain nearly uniform A/B/C/D answer proportions, while Tasks 3 and 5 contain balanced binary labels; Task1 is excluded because it is a residue-level prediction task without fixed answer options. The resulting distributions are shown in Figure~\ref{fig:answer-distribution}.

\section{Prompt Templates and Qualitative Examples}

\subsection{Prompt Templates}

EpiBench uses a consistent prompt interface across tasks: a task-specific biological role, a closed-book sequence input, a concise task question, and a deterministic JSON answer format. The explicit-reasoning setting used in the main experiments asks models to expose intermediate reasoning before the final JSON line; the direct-answer setting in Section~4.3 keeps the same task definition and output schema but replaces the reasoning instruction with ``Respond with ONLY a JSON object and no other text.''

\begin{systempromptbox}{System Prompt with Explicit Reasoning (CoT)}
\textbf{Role}: You are an expert in epitope-centered antibody drug discovery. Your task is to answer the question by analyzing only the provided antigen and antibody sequence information.
\par\smallskip
\textbf{Question}: \promptplaceholder{question}
\par\smallskip
\textbf{Sequence evidence}: \promptplaceholder{sequence}
\par\smallskip
\textbf{Choices}: \promptplaceholder{choices}
\par\smallskip
\textbf{Instruction}: Think step-by-step using biological sequence evidence. Do not copy or restate long sequences in the response; quote only short substrings if needed.
\par\smallskip
\textbf{Output format}: End with a JSON object and no extra text after it: \texttt{\{"answer": "}\promptplaceholder{answer}\texttt{"\}}.
\end{systempromptbox}

\begin{systempromptbox}{System Prompt without Explicit Reasoning (Non-CoT)}
\textbf{Role}: You are an expert in epitope-centered antibody drug discovery. Your task is to answer the question by analyzing only the provided antigen and antibody sequence information.
\par\smallskip
\textbf{Question}: \promptplaceholder{question}
\par\smallskip
\textbf{Sequence evidence}: \promptplaceholder{sequence}
\par\smallskip
\textbf{Choices}: \promptplaceholder{choices}
\par\smallskip
\textbf{Instruction}: Provide the final answer directly. Do not include intermediate reasoning or explanatory text.
\par\smallskip
\textbf{Output format}: Output only a JSON object: \texttt{\{"answer": "}\promptplaceholder{answer}\texttt{"\}}.
\end{systempromptbox}

\subsection{Task-specific Prompt Examples}

The examples below are shortened excerpts derived from the rendered EpiBench CoT user prompts after de-identification. Except for truncating long antigen or antibody sequences and omitting residue-number columns for readability, the field names, candidate lists, and question wording follow the prompts used in evaluation.

\begin{examplebox}{Example: Targetable Region Discovery}
\textbf{Antigen (1310 residues; numbering omitted)}:
\par\smallskip
\texttt{MFVFLVLLPLVSSQCVNLTTRTQLPPAYTN...SHPQFEK}
\par\smallskip
\textbf{Question}: Predict AT MOST 50 residue POSITIONS on this antigen most likely to be part of an antibody epitope (1-based positions, ranked most-likely first; no more than 50). Reason briefly, then give your final answer in the required JSON format with all positions in the array.
\end{examplebox}

\begin{examplebox}{Example: Antibody-conditioned Epitope Identification}
\textbf{Antibody heavy chain (VH, variable domain)}:
\par\smallskip
\texttt{QVQLVQSGAEVKKPGSSVKVSCKASGGTFSSYP...CARGIASAGTPDYFFY} \\
\hspace*{1em}CDR1: GGTFSSYP \quad CDR2: IIPLFGTA \quad CDR3: ARGIASAGTPDYFFY
\par\smallskip
\textbf{Antibody light chain (VL, variable domain)}:
\par\smallskip
\texttt{EIVLTQSPGTLSLSPGERATLSCRASQSISSSY...CQQYGSSPYT} \\
\hspace*{1em}CDR1: QSISSSY \quad CDR2: GAS \quad CDR3: QQYGSSPYT
\par\smallskip
\textbf{Antigen (894 residues; numbering omitted)}:
\par\smallskip
\texttt{TQVCTGTDMKLRLPASPETH...KIEWHESRHHHHH}
\par\smallskip
\textbf{Candidate epitopes (each token = residue+position on the antigen)}: \par
(A) P557, E558, D560, Q561, K569, D570, P571, P572, F573, P579, Y588, M589, I591, K593, D596, A600, Q602, P603, C604, P605 \\
(B) E373, Q376, E379, D395, S396, L397, P398, D399, S401, V402, Q404, G427, S429, W430, R434, T450, H451, R477, E481, V483, E485, G486, L487, P501, G502, T504, Q505 \\
(C) L654, G655, G656, P657, S686, H687, E688, D689, P690, E691, K741, N744, K745, A746, L747, P748, A749, P750 \\
(D) T669, L670, M671, I672, S673, H729, Q730, L733, E801, G804, V841, S843, M847, A850, L851, H852, N853, H854, Y855, T856, Q857, K858, S859, L860, S861
\par\smallskip
\textbf{Question}: Which candidate epitope does THIS antibody bind? Reason step-by-step, then give your final answer in the required JSON format.
\end{examplebox}

\begin{examplebox}{Example: Epitope Binning}
\textbf{Antigen (365 residues; numbering omitted)}:
\par\smallskip
\texttt{DNLWVTVYYGVPVWKDADTT...NWRSELYKYKVVQIE}
\par\smallskip
\textbf{Antibody A --- heavy chain (H)}: \texttt{HVQLVQSGGGVKKIGAAVRISCEVSGYNFMDQF...ARGPSGENYPFHY} \\
\hspace*{1em}CDR1: GYNFMDQF \quad CDR2: MNPIYGQV \quad CDR3: ARGPSGENYPFHY
\par\smallskip
\textbf{Antibody A --- light chain (L)}: \texttt{LTQPASMSASPGQSVTISCSGTRHIIS...YICNTYEF} \\
\hspace*{1em}CDR1: RHIIS \quad CDR2: DDD \quad CDR3: NTYEF
\par\smallskip
\textbf{Antibody B --- heavy chain (H)}: \texttt{QVQLVQSGGQMKKPGESMRISCRASGYEFIDCT...TRGKNCDYNWDFEH} \\
\hspace*{1em}CDR1: GYEFIDCT \quad CDR2: LKPRGGAV \quad CDR3: TRGKNCDYNWDFEH
\par\smallskip
\textbf{Antibody B --- light chain (L)}: \texttt{VLTQSPGTLSLSPGETAIISCRTSQYGS...CQQYEF} \\
\hspace*{1em}CDR1: QYGS \quad CDR2: SGS \quad CDR3: QQYEF
\par\smallskip
\textbf{Question}: Do antibody A and antibody B bind overlapping epitope regions on this antigen? Reason step-by-step, then give your final answer in the required JSON format.
\end{examplebox}

\begin{examplebox}{Example: Functional Epitope Assessment}
\textbf{Antigen sequence (824 residues; numbering omitted)}:
\par\smallskip
\texttt{MEFIPTQTFYNRRYQPRPWT...RTRNSDPEHGGSTV}
\par\smallskip
\textbf{Desired antibody-related functional outcome}: a functional antibody response (neutralization, inhibition of activity, protection, or antibody-mediated cytotoxicity)
\par\smallskip
\textbf{Candidate B-cell epitopes (peptides from this antigen)}: \par
A. residues 567-584: VPRNAELGDRKGKIHIPF \\
B. residues 326-335: STKDNFNVYK \\
C. residues 12-21: RRYQPRPWTP \\
D. residues 300-309: EDNVMRPGYY
\par\smallskip
\textbf{Question}: Which candidate epitope is experimentally associated with the desired functional antibody response? Reason step-by-step, then give your final answer in the required JSON format.
\end{examplebox}

\begin{examplebox}{Example: Escape Assessment}
\textbf{Antibody heavy chain (VH, variable domain)}:
\par\smallskip
\texttt{EVQLVQSGAEVKKPGASVKVSCKASGYTFTGYY...CARAAPFYDFWSGYSYFDY} \\
\hspace*{1em}CDR1: GYTFTGYY \quad CDR2: INPISSGT \quad CDR3: ARAAPFYDFWSGYSYFDY
\par\smallskip
\textbf{Antibody light chain (VL, variable domain)}:
\par\smallskip
\texttt{EIVMMQSPGTLSLSPGERATLSCRASQSVSSSY...CQQYGSSAWT} \\
\hspace*{1em}CDR1: QSVSSSY \quad CDR2: GAS \quad CDR3: QQYGSSAWT
\par\smallskip
\textbf{Wild-type antigen (201 residues; numbering omitted)}:
\par\smallskip
\texttt{NITNLCPFGEVFNATRFASV...PLQSYGFQPTNGVGYQPYRV...PATVCGPKKST}
\par\smallskip
\textbf{Mutation}: \par
position 176: Q\,$\rightarrow$\,A
\par\smallskip
\textbf{Question}: Will this antigen mutation cause the antibody to lose binding? Reason step-by-step, then give your final answer in the required JSON format.
\end{examplebox}

\subsection{Representative Error Case}

The case below shows a real incorrect Task5 response from DeepSeek-V4-Pro. We do not repeat the task prompt; instead, we show the stored model trace in layers, with only long sequence-counting spans shortened by ellipses. The source record contains a local RBD mutation at position 154 that corresponds to full-spike E484H, and the hidden gold label is \texttt{escape}.

\begin{casebox}{Case Study: Coordinate Misalignment in Escape Assessment}
\textbf{Sample}: Task5, \texttt{task5\_0161} \qquad
\textbf{Model}: DeepSeek-V4-Pro \qquad
\textbf{Parsed answer}: \texttt{non\_escape} \qquad
\textbf{Gold label}: \texttt{escape}
\par\medskip

\textbf{Layer 1: Antigen recognition.}
\emph{Model output}: ``The antigen is presumably the SARS-CoV-2 spike protein receptor-binding domain (RBD) based on the sequence? Let's check: The sequence provided is 201 residues, starting with \texttt{NITNLCPFGE...}. This matches the SARS-CoV-2 RBD sequence.''
\par\smallskip
\textit{Summary}: The model correctly recognizes the biological context as a SARS-CoV-2 RBD-derived sequence.
\par\medskip

\textbf{Layer 2: Local-to-global coordinate mapping.}
\emph{Model output}: ``Position 154 is indeed E (glutamic acid). Mutation E154H. \ldots{} \errorhl{position 154 in this numbering corresponds to full spike position 472}? Actually: 1\,$\rightarrow$\,319, 2\,$\rightarrow$\,320, \ldots{} so n\,$\rightarrow$\,318+n. So \errorhl{154\,$\rightarrow$\,318+154 = 472}.''
\par\smallskip
\textit{Summary}: The failure begins when the model assumes an incorrect RBD offset and converts the local mutation into \errorhl{\texttt{E472H}} rather than the source-record \texttt{E484H}.
\par\medskip

\textbf{Layer 3: Speculative antibody identification.}
\emph{Model output}: ``The heavy chain CDR3 sequence \texttt{ARDAGKSPLSYWYFDL} is quite distinctive. \ldots{} Another possibility: This might be the antibody `C105' or `C144'? \ldots{} I think the antibody `C135' (or `C144') might have a similar CDR3.''
\par\smallskip
\textit{Summary}: The model then uses memorized antibody associations as supporting evidence, although the evaluation input is de-identified and the source metadata does not provide these names to the model.
\par\medskip

\textbf{Layer 4: Escape decision from the wrong site.}
\emph{Model output}: ``I recall the C144 escape map. \ldots{} The escape mutations are at E484, F490, and maybe S494. \errorhl{I don't think E472 is a hit}. \ldots{} Therefore, the answer should be \errorhl{\texttt{non\_escape}}.''
\par\smallskip
\textit{Summary}: The model recalls that E484 can be escape-associated, but because it has already mapped the mutation to E472, it reaches the opposite decision.
\par\medskip

\textbf{Evaluation}: The output is parsed deterministically as \texttt{\{"answer": "non\_escape"\}} and compared against the gold label \texttt{\{"answer": "escape"\}}.
\par\smallskip
\textbf{Error cause}: An incorrect local-to-global coordinate mapping leads the model to evaluate \texttt{E472H} instead of the true \texttt{E484H} mutation.
\end{casebox}

This example illustrates why EpiBench evaluates sequence-grounded answers rather than only the plausibility of generated explanations. The response contains domain-relevant biological knowledge, but the final answer is determined by a concrete sequence-coordinate error.

\clearpage

\section{Supplementary Experimental Analyses}

\subsection{Performance across Task Subcategories}

\begin{figure}[H]
    \centering
    \includegraphics[width=\linewidth]{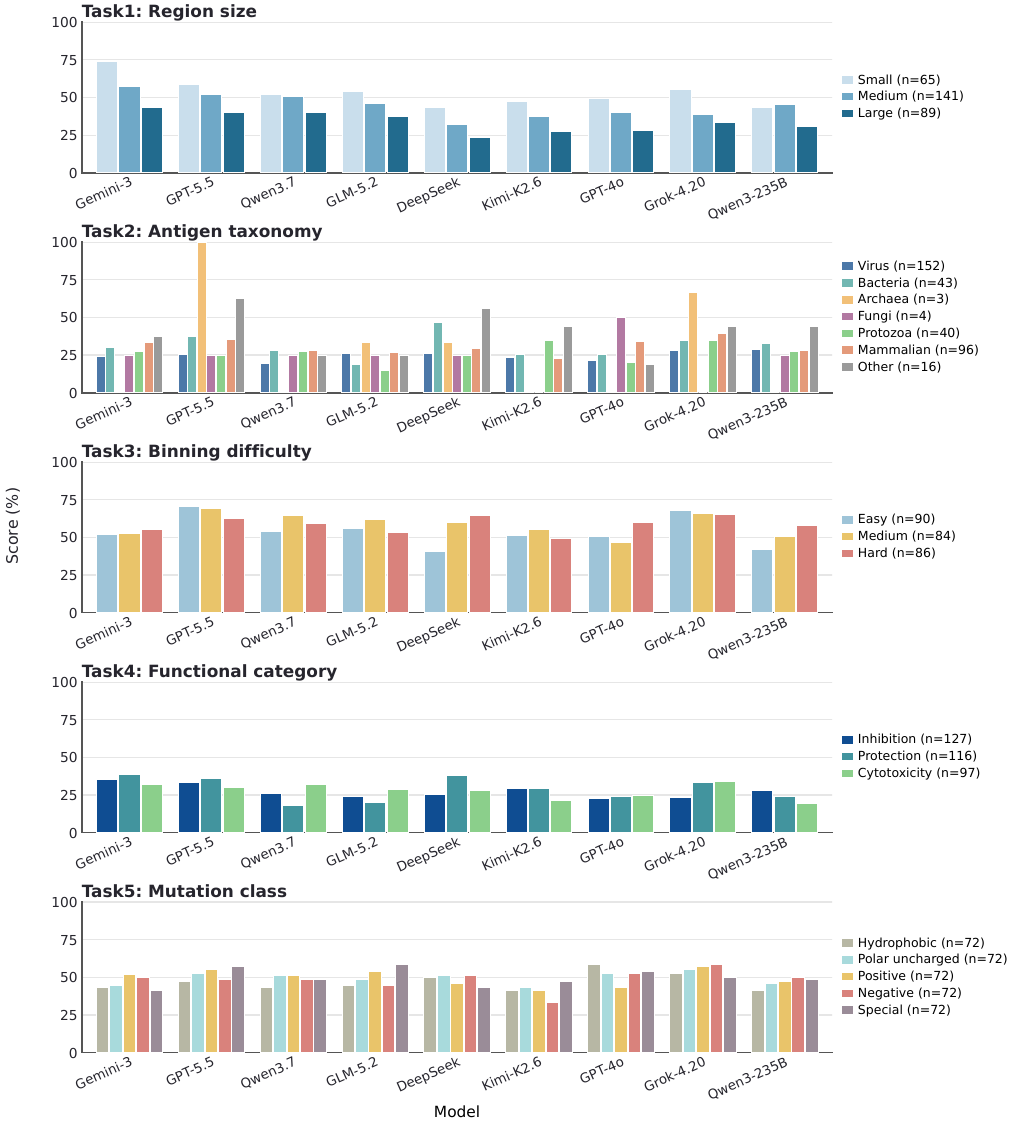}
    \caption{Subcategory-wise model performance across EpiBench. Each panel corresponds to one task, the x-axis lists evaluated LLMs, and grouped bars denote the task-specific main subcategories. Scores are reported using the same task metrics as the main evaluation and shown as percentages. Task4 is summarized by primary functional categories. Rare Task2 taxonomy categories such as Archaea and Fungi contain few examples and should be interpreted cautiously.}
    \label{fig:subcategory-grouped-bars}
\end{figure}

EpiBench contains controlled heterogeneity within each task, including target-region size, antigen taxonomy, binning difficulty, functional mechanism, and mutation chemistry. We therefore analyze whether the overall results are driven by a narrow subset of examples or persist across the main task-defined subcategories. Figure~\ref{fig:subcategory-grouped-bars} reports model performance across these subcategories while keeping the model order fixed.

The subcategory trends provide complementary evidence to the aggregate results in Section~4. Task1 shows a consistent decrease from small to large target regions for most models, suggesting that residue-level region discovery becomes harder as the target set becomes more complex. Task2 varies across antigen taxonomy, although rare categories should not be over-interpreted. Task3 does not follow a simple monotonic pattern over the predefined difficulty groups, indicating that epitope binning depends on more than this coarse difficulty label. Task4 remains low across inhibition, protection, and cytotoxicity, and Task5 is comparatively similar across mutation chemistry classes. Overall, the subcategory analysis suggests that EpiBench difficulty is not caused by a single dominant subset, but by task-specific forms of sequence-grounded reasoning.

\subsection{Answer-position Balance and Option-wise Accuracy}

For tasks with discrete outputs, we examine whether model performance is affected by answer-position or label preference rather than the biological content of the prompt. EpiBench therefore balances the gold answer distribution for multiple-choice and binary tasks, and we further compute option-wise accuracy by aggregating the nine main LLM runs after deduplicating predictions by final test sample ID.

\begin{figure}[H]
    \centering
    \includegraphics[width=0.98\linewidth]{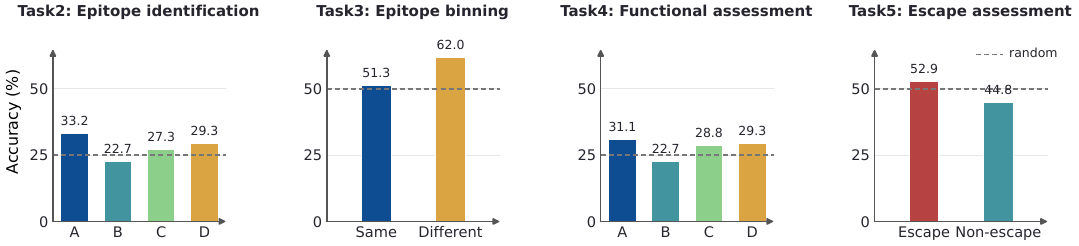}
    \caption{Option-wise accuracy under balanced answer distributions. Bars report aggregated accuracy over the nine evaluated LLMs for each gold answer option or binary label, and dashed lines denote random baselines. The results show no strong systematic preference for a specific answer position at the aggregate level, supporting the fairness of the discrete-choice evaluation.}
    \label{fig:option-wise-accuracy}
\end{figure}

The option-wise results are broadly balanced across answer positions, especially for the four-way multiple-choice tasks. This indicates that the reported scores are not explained by a simple shortcut such as consistently favoring one option letter or one binary label. The remaining variation is therefore better interpreted as model behavior on the underlying sequence-based decision, rather than an artifact of an imbalanced answer key.

\end{document}